\documentclass{article}

\usepackage[eandd, preprint]{neurips_2026}

\usepackage[utf8]{inputenc} 
\usepackage[T1]{fontenc}    
\usepackage{hyperref}       
\usepackage{url}            
\usepackage{booktabs}       
\usepackage{multirow}       
\usepackage{graphicx}       
\usepackage{amsfonts}       
\usepackage{nicefrac}       
\usepackage{microtype}      
\usepackage{xcolor}         
\usepackage{placeins}       
\usepackage{float}          
\usepackage{caption}        
\title{SceneForge: Structured World Supervision from 3D Interventions}

\author{
\begin{tabular}{cc}
Jizhizi Li$^{1}$\thanks{Corresponding author} & Jiayang Ao$^{1}$ \\
\texttt{jizhizili@canva.com} & \texttt{jiayangao@canva.com} \\
Danny Wicks$^{1}$ & Petru-Daniel Tudosiu$^{1}$ \\
\texttt{dannywicks@canva.com} & \texttt{ptudosiu@canva.com} \\
\multicolumn{2}{c}{$^{1}$Canva Research}
\end{tabular}
}

\begin{document}

\maketitle

\begin{abstract}
  Many multimodal learning tasks require supervision that remains consistent across edits, viewpoints, and scene-level interventions. However, such supervision is difficult to obtain from observation-level datasets, which do not expose the underlying scene state or how changes propagate through it. We present \textbf{SceneForge}, an intervention-driven framework that generates structured supervision from editable 3D world states. SceneForge represents each scene as a persistent world with semantic, geometric, and physical dependencies. By applying explicit interventions (e.g., object removal or camera variation) and propagating their effects through scene dependencies, SceneForge renders supervision that remains consistent with object structure and scene-level effects. This produces aligned outputs including counterfactual observations, multi-view observations, and effect-aware signals such as shadows and reflections, all derived from a shared world state rather than post hoc image-space processing. We instantiate SceneForge using Infinigen and Blender to construct a licensing-clean indoor supervision resource with a large number of counterfactual pairs and aligned annotations from over 2K scenes, covering both diverse single-view and registered multi-view settings. Under matched training budgets, incorporating SceneForge supervision improves both object removal and scene removal performance across multiple benchmarks in both quantitative and qualitative evaluation. These results indicate that modeling supervision as structured state transitions in editable worlds provides a practical and scalable foundation for intervention-consistent multimodal learning.
\end{abstract}

\section{Introduction}

Modern multimodal systems increasingly require supervision that is not only large-scale, but also structured, controllable, and aligned across edits, viewpoints, and modalities. In many practical settings, models must reason jointly over text, images, videos, geometric cues, and object-level structure while remaining consistent under scene edits, temporal evolution, and camera variation. Yet this kind of supervision is difficult to obtain from observation-level datasets, which expose rendered snapshots rather than the underlying scene state and therefore rarely provide matched interventions, physically consistent counterfactuals, layered decompositions, relation-aware annotations, or aligned multimodal signals for the same scene. Diagnostic synthetic datasets such as CLEVR \citep{johnson2017clevr} have demonstrated the value of controllable world generation for visual reasoning, but they remain focused on fixed reasoning tasks rather than intervention-driven supervision for modern generative and editing models.

Existing efforts partially address this gap from two directions. Real or hybrid editing datasets provide task-relevant examples for image and video manipulation, including manually annotated and automatically generated instruction-based editing resources, but they typically construct supervision at the observation level, for example as before-and-after pairs, task-specific edits, or localized annotations \citep{chen2023magicbrush,kim2025orida,an2024ultraedit,li2022bridgingcomposite}. Synthetic 3D pipelines, in contrast, provide strong scene control and rich render-time annotations through controllable simulators and procedural generators \citep{kar2019metasim,greff2022kubric,raistrick2023infinite}, but they often treat scenes primarily as configurations to be rendered rather than as persistent worlds whose state can be manipulated, queried, and updated in a structured manner. What remains missing is a framework that models supervision generation as structured state transition in an editable world, so that a single intervention can induce consistent updates across entities, views, modalities, and scene-level effects.

In this work, we present \textbf{SceneForge}, a world-centric framework for structured supervision from 3D interventions. The key idea is to cast supervision generation as intervention on an editable world state rather than as post hoc construction from observations. SceneForge represents each scene as a structured world composed of stable structural elements and editable dynamic objects. It models semantic, geometric, and physical dependencies among these entities, and propagates the effects of interventions through the maintained world so that changes remain consistent with occlusion, support, visibility, lighting, and other scene-level constraints.

Under this formulation, each world state and state transition induces aligned \textbf{observation- and layer-level} supervision, including RGB observations, multi-view renderings, counterfactual edits, decomposition outputs, geometry-aware signals, and effect-aware channels such as shadows and reflections. Because these outputs are derived from the \textbf{same} maintained scene graph, they are aligned by construction rather than matched after the fact in image space. This same formulation also provides a natural basis for extending SceneForge toward richer \emph{linguistic} supervision and broader asset diversity.

We evaluate SceneForge through dataset instantiation and statistics, compositional layer decomposition, intervention-driven counterfactual generation, and downstream applications on both object-level and scene-level removal tasks.

Our contributions are fourfold. First, we introduce SceneForge, a world-centric framework in which structured multimodal supervision is generated through explicit interventions and dependency-aware state propagation. Second, we instantiate and release SceneForge with Infinigen and Blender as a licensing-clean framework and resource, including aligned object-removal datasets with complementary coverage-oriented single-view and multi-view splits together with the supporting code and documentation. Third, we characterize these datasets through statistics, compositional layer decomposition, and intervention-driven counterfactual analyses grounded in cross-view and cross-edit consistency. Fourth, we present downstream applications on both object-level and scene-level removal tasks under matched training settings. More broadly, the same formulation naturally extends to additional supervision types, such as dense language or point clouds, and is not restricted to Infinigen-native assets.

\section{Related Work}

\textbf{Editing datasets and hybrid supervision pipelines.}
Recent work has produced increasingly rich resources for instruction-guided and object-centric image editing, including instruction-following editors, manually curated datasets, automatically generated supervision, and hybrid pipelines that combine real imagery with controlled editing operations \citep{brooks2023instructpix2pix,chen2023magicbrush,liu2024referringedit,kim2025orida,an2024ultraedit}. Related work on image matting and object-aware image manipulation further emphasizes fine-grained foreground understanding and decomposition-aware supervision \citep{li2022bridgingcomposite}. These resources provide task-relevant supervision for image or video editing, but they typically construct labels at the observation level, such as paired edits, localized annotations, or instruction-conditioned targets. Concurrent work such as ROSE \citep{miao2025roseremoveobjectseffects} focuses on side-effect-aware object removal in videos, but remains centered on paired counterfactual video supervision in more limited scene settings. In contrast, SceneForge treats supervision as the outcome of explicit interventions on an underlying world state, enabling broader editable functions, freer camera control, and aligned multimodal outputs in addition to edited observations, decompositions, relations, and cross-view or temporal consistency.

\textbf{Synthetic 3D data generation.}
A complementary line of work studies synthetic data generation through controllable simulators, synthetic benchmarks, procedural generators, and rendering pipelines \citep{richter2016playing,ros2016synthia,dosovitskiy2017carla,qiu2017unrealcv,kar2019metasim,denninger2019blenderproc,greff2022kubric,raistrick2023infinite}. Related embodied and interactive environments further illustrate the value of procedural or world-based simulation for controllable learning settings \citep{gan2020threedworld,savva2019habitat,deitke2022procthor}. These systems provide flexible scene control, scalable rendering, and rich annotations, but they often expose scenes primarily as rendering configurations rather than as persistent structured worlds whose interventions induce consistent updates across entities, modalities, and views. SceneForge builds on the strengths of synthetic generation while centering the editable world state and its transitions as the primary object of supervision.

\textbf{Controllable worlds, reasoning datasets, and structured scene representations.}
A broader literature has shown the value of structured scene control for visual reasoning and relational understanding. Diagnostic synthetic datasets such as CLEVR use controlled worlds to probe compositional reasoning \citep{johnson2017clevr}, while Visual Genome, GQA, visual relationship detection, scene graph modeling, and AGQA provide relation-rich supervision for scene understanding and grounded reasoning \citep{krishna2017visualgenome,hudson2019gqa,lu2016vrd,zellers2018neuralmotifs,grunde2021agqa}. SceneForge is motivated by the same appreciation for structured world representations, but uses structure not mainly for reasoning benchmarks or relation annotation, but to generate aligned multimodal supervision under explicit interventions.

\textbf{Amodal, layered, and decomposition-aware supervision.}
Another related line of work studies supervision beyond directly visible image evidence, including amodal segmentation, amodal appearance completion, matting, and layered scene understanding \citep{li2016amodal,zhu2017semanticamodal,qi2019kins,li2020oneshotamodal,ao2025openworldamodal,li2021deepmatting,li2023referringmatting}. These works highlight the importance of reasoning about occluded content, foreground structure, and decomposition-aware supervision, but they typically focus on specific annotation targets. SceneForge instead treats such signals as one component within a broader generation framework that jointly produces edited observations, decomposition-level outputs, and relation-aware annotations from the same world state.

\section{SceneForge Framework}
\label{sec:framework}

Figure~\ref{fig:sceneforge-pipeline} summarizes the practical pipeline studied in this work: editable 3D scenes are first acquired or generated, abstracted into persistent world states, modified through structured interventions and scene querying, rendered into aligned supervision, and finally assembled into task-ready datasets. Section~\ref{sec:framework} follows this same flow.

\subsection{World State Representation}

\begin{figure*}[t]
    \centering
    \captionsetup{skip=7pt}
    \includegraphics[width=\textwidth]{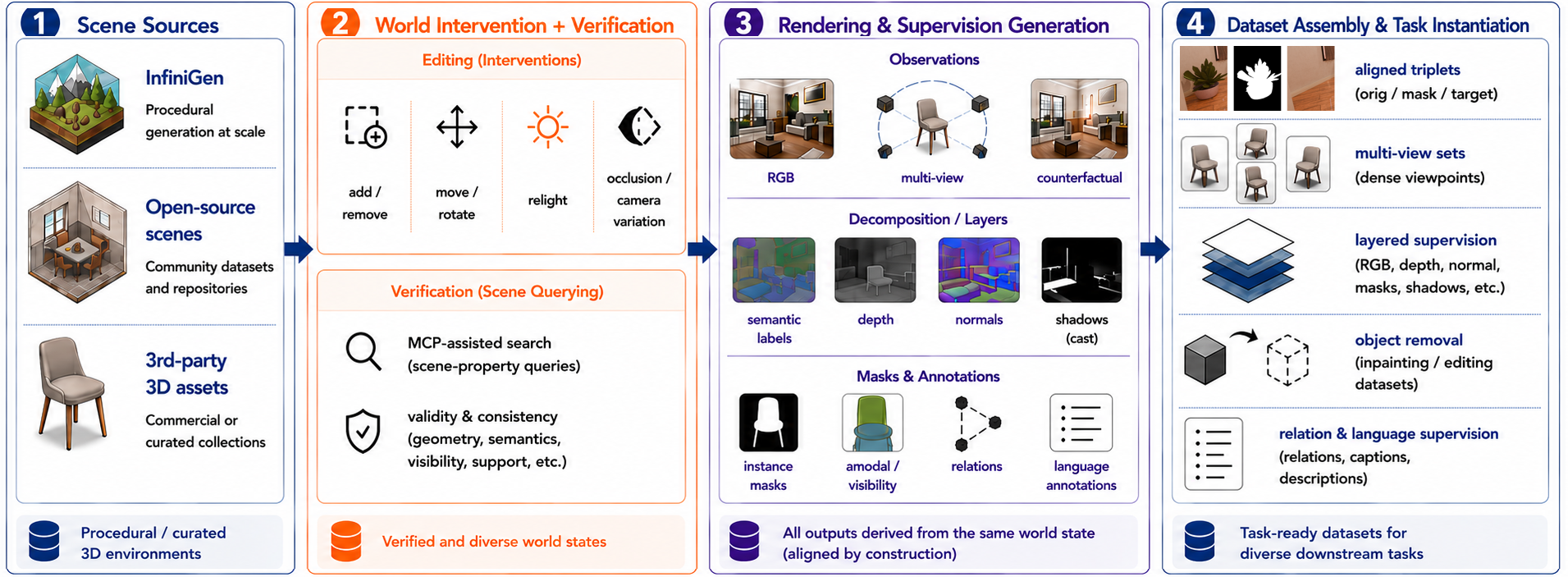}
    \caption{Overview of the SceneForge pipeline. Editable 3D scenes are first acquired or generated from procedural or third-party sources, then modified through structured world interventions and verified through scene-querying modules. The resulting world states are rendered into aligned observations, decomposition layers, and annotations, which are finally assembled into task-specific supervision resources such as aligned triplets and multi-view datasets.}
    \label{fig:sceneforge-pipeline}
\end{figure*}

The first step in Figure~\ref{fig:sceneforge-pipeline} converts editable scenes from procedural or third-party sources into persistent world states rather than flat rendering configurations. We represent a scene at time $t$ as
\[
w_t = (\mathcal{S}_t, \mathcal{D}_t, \mathcal{A}_t, \mathcal{R}_t, \mathcal{E}_t),
\]
where $\mathcal{S}_t$ denotes \textbf{structural elements}, $\mathcal{D}_t$ denotes \textbf{dynamic objects}, $\mathcal{A}_t$ their attributes, $\mathcal{R}_t$ their structured relations, and $\mathcal{E}_t$ the environmental factors required for rendering and supervision generation. \textbf{Structural elements} include stable scene components such as floors, walls, large support surfaces, or fixed background geometry. \textbf{Dynamic objects} are the entities that may be inserted, removed, moved, replaced, or otherwise modified by interventions. This separation defines what is editable, what acts as a stable anchor, and which changes must be propagated when the world is modified.

Attributes capture semantic, geometric, and physical state, including category, instance identity, pose, extent, occupancy, material, support status, contact, and lighting-related properties when relevant. The relation set $\mathcal{R}_t$ stores dependencies among entities, including support, attachment, containment, relative position, visibility, and occlusion. Together these variables allow SceneForge to treat a scene as a structured system rather than a collection of independently rendered objects.

This representation is persistent across edits. Instead of generating each output independently, SceneForge maintains a state before intervention, applies an explicit modification, and then updates the resulting world into a new consistent state. In this way, all supervision signals are tied to the same underlying world and can be aligned across modalities, viewpoints, and time.

\subsection{Intervention Model}

Given a world state $w_t$, SceneForge defines an intervention as an explicit operation $\iota$ that acts on one or more dynamic objects or environment variables. In the pipeline of Figure~\ref{fig:sceneforge-pipeline}, this is the stage where scenes are modified and verified through scene-querying modules before rendering. The role of the intervention model is to specify which changes are permitted, which entities they may affect directly, and what parameters are required to instantiate the edit. Typical interventions include object insertion, object removal, object relocation, pose adjustment, lighting modification, occlusion manipulation, and controlled camera variation. We write the resulting provisional edit as
\[
\tilde{w}_{t+1} = \mathcal{I}(w_t, \iota).
\]

The intervention model is constrained by scene validity. Not all edits are meaningful or physically plausible. For example, a structural wall should not be treated as a freely movable object, and an inserted object should be placed in a location that is geometrically feasible and semantically coherent with the scene. Similarly, a relocation should preserve basic scene plausibility with respect to support surfaces, collisions, and task-specific constraints. The scene-querying stage in Figure~\ref{fig:sceneforge-pipeline} operationalizes these checks by verifying edit targets, affected entities, and view conditions before supervision is emitted. These validity conditions ensure that interventions correspond to interpretable world edits rather than arbitrary rendering perturbations.

This formulation distinguishes SceneForge from pipelines that construct supervision directly as observation pairs. In SceneForge, the edit is defined first at the world level, with explicit semantics and constraints. Rendered outputs are then derived from the resulting world states, rather than serving as the primary object being edited.

\subsection{Dependency-Aware State Propagation}

A local intervention typically has non-local consequences. Removing an object changes not only its own visibility, but also the appearance of newly revealed background regions, occlusion relationships, support interactions, and potentially shadows or reflections. Moving an object changes its relative position to other entities, alters visibility from multiple viewpoints, and may modify contact or containment relations. Changing illumination affects not only image appearance, but also decomposition signals such as shadow-sensitive or reflectance-related outputs. SceneForge captures these secondary effects through a dependency-aware state propagation step.

Formally, after an intervention produces a provisional state $\tilde{w}_{t+1}$, SceneForge applies a propagation operator $\mathcal{P}$ to obtain a consistent updated state $w_{t+1} = \mathcal{P}(\tilde{w}_{t+1})$. The operator updates all affected attributes and relations that depend on the edited entities. In practice, this includes geometric updates, visibility and occlusion recomputation, relation maintenance, support consistency, and lighting-aware adjustments when the intervention changes scene illumination or material interaction. The exact propagation rules may vary across scenes and assets, but the principle is fixed: dependent scene state must be updated before supervision is emitted.

This propagation stage is the core difference between SceneForge and naive render-pair generation. A naive approach may generate before-and-after images without explicitly modeling how the edit changes the underlying world. SceneForge instead treats an intervention as a structured state transition, ensuring that all downstream supervision signals remain semantically and physically consistent with the edited world.

\subsection{Multimodal Supervision Generation}

Once a consistent world state or transition is available, SceneForge renders aligned supervision across multiple modalities from the same scene representation. In Figure~\ref{fig:sceneforge-pipeline}, this stage converts edited worlds into observations, decomposition layers, text labels, and annotations. For an output type $m \in \mathcal{M}$ and camera or view configuration $c$, we write
\[
y_m = \mathcal{G}_m(w_t, w_{t+1}, c), \qquad m \in \mathcal{M},
\]
where $\mathcal{M}$ may include RGB observations, multi-view renderings, counterfactual edits, decomposition outputs, geometry-aware signals, semantic or instance annotations, amodal targets, language-grounded descriptions, and relation annotations.

The central property of this design is alignment by construction. Because these outputs are derived from the same world state, they share object identities, scene geometry, camera configuration, and intervention history. This makes it possible to align supervision across pixels, frames, views, and modalities without post hoc matching. The same intervention that produces an edited image can also produce the corresponding amodal mask, relation annotation, decomposition target, or multi-view observation under consistent scene assumptions.

This design also supports multiple supervision granularities, including single-state supervision, before-and-after pairs, and multi-step sequences for evolving scenes.

\subsection{Dataset Assembly and Task Instantiation}

The final stage of SceneForge assembles rendered outputs into task-ready training data by selecting emitted signals, intervention metadata, and task-specific packaging. For example, an object-removal sample may pair an original image with an edited target, masks, and decomposition targets; an amodal sample may pair a rendered observation with amodal and visibility-aware annotations; and a multi-view sample may expose several projections of the same world state.

This last block of Figure~\ref{fig:sceneforge-pipeline} packages aligned outputs into concrete supervision resources such as triplets and multi-view datasets. Different downstream tasks therefore consume different projections of the same underlying world process; we return to their concrete dataset instantiation and statistics in Section~\ref{sec:dataset-stats}.

\section{Empirical Study}

We evaluate SceneForge from four perspectives: dataset statistics, compositional layer decomposition, intervention-driven counterfactual generation, and downstream removal applications.

\subsection{Dataset Instantiation and Statistics}
\label{sec:dataset-stats}

We organize SceneForge into two raw corpora and several derived counterfactual datasets. For clean open release, the current instantiation is built primarily from \textbf{Infinigen} \citep{raistrick2023infinite} and editable \texttt{.blend} scenes, while remaining agnostic to the asset source when redistribution permits. We use Infinigen indoor scenes as the running example here, but the framework is not limited to indoor procedural content. SceneForge produces raw object-linked supervision corpora and downstream datasets for object removal, multi-view re-editing, and structural-only scene removal; implementation details are deferred to the supplementary material. \textbf{SceneForge (S)} is the broader single-camera corpus, with $2{,}369$ rooms, equivalently $2{,}369$ single-view scene instances, aligned depth maps, instance segmentation, surface normals, language labels, and $25{,}688$ object-linked layers. These layers may represent dependency-aware groupings and include amodal completion when occluded objects can be recovered from the maintained world state. \textbf{SceneForge (M)} is the denser multi-camera corpus, with $517$ rooms and $2{,}772$ registered multi-view scene instances, the same annotation types, and $34{,}865$ object-linked image layers under $8$ camera settings.

\begin{table}[H]
  \vspace{-0.3em}
  \setlength{\abovecaptionskip}{2pt}
  \setlength{\belowcaptionskip}{0pt}
  \caption{SceneForge data distribution.}
  \label{tab:dataset_stats}
  \begin{center}
    \tiny
    \setlength{\tabcolsep}{2pt}
    \renewcommand{\arraystretch}{0.88}
    \resizebox{0.98\columnwidth}{!}{%
    \begin{tabular}{lrrrrl}
      \toprule
      Corpus & Unique rooms & Unique scenes & Images / layers & Items / scene & Distribution role \\
      \midrule
      SceneForge (S, raw layers) & 2,369 & 2,369 & 25,688 & 10.84 & single-view object-linked supervision \\
      SceneForge (M, raw layers) & 517 & 2,772 & 34,865 & 12.58 & multi-view object-linked supervision \\
      \midrule
      SceneForge-Removal (S, $>0.3\%$ mask) & 2,358 & 13,419 & 13,419 & 1.00 & filtered single-view object-removal instances \\
      SceneForge-Removal (M, $>0.3\%$ mask) & 330 & 3,999 & 3,999 & 1.00 & filtered multi-view object-removal instances \\
      SceneForge-IndoorDynamicRemoval & 2,435 & 2,435 & 2,435 & 1.00 & scene-level structural-only removal \\
      \bottomrule
    \end{tabular}}
  \end{center}
  \vspace{-0.8em}
\end{table}

\begin{figure}[H]
    \centering
    \includegraphics[width=0.90\linewidth]{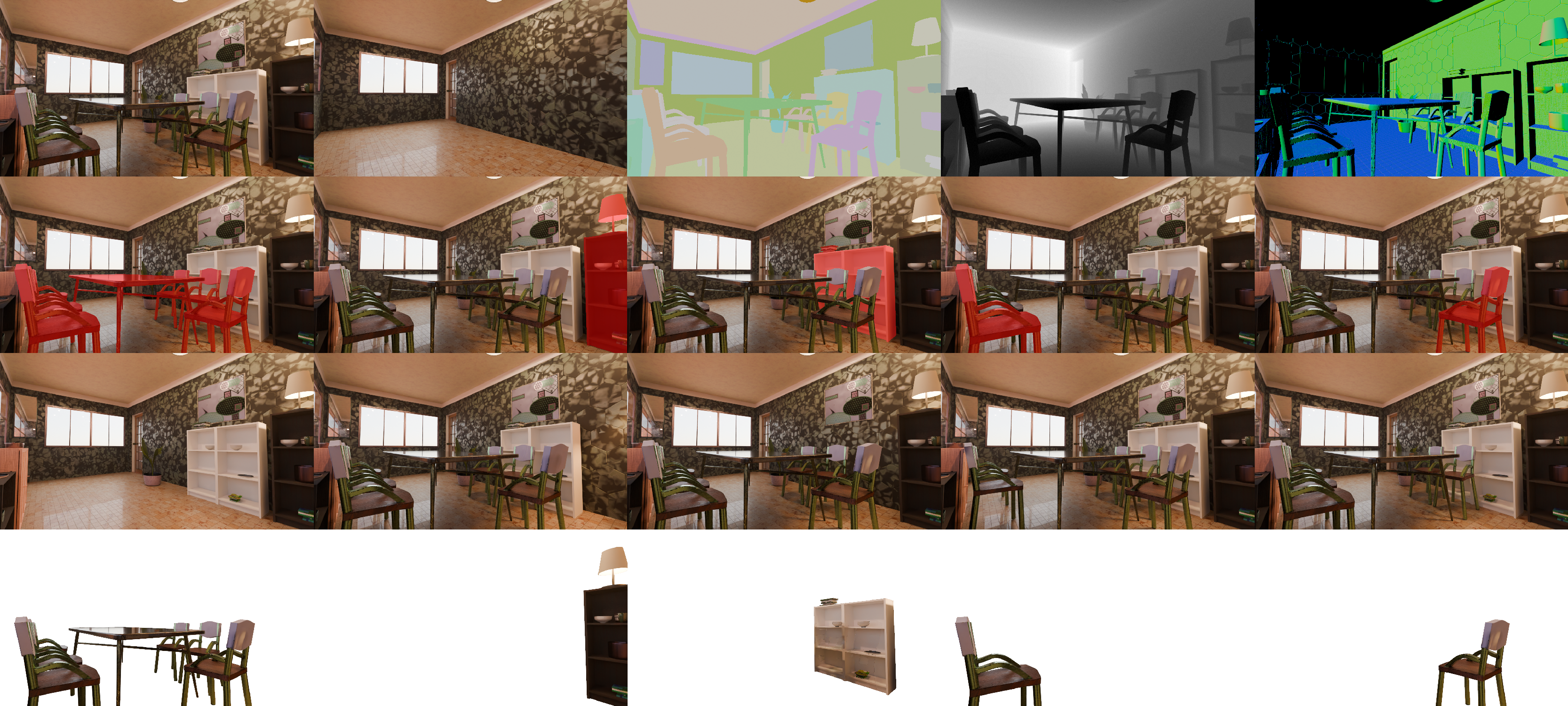}
    \caption{Object-linked layer decomposition in an Infinigen scene. From one scene state, SceneForge exposes aligned object, effect, and component layers beyond a single fixed edit pair. Only five representative layers are shown due to page limits.}
    \label{fig:infinigen-layer-demo}
\end{figure}

To derive downstream datasets, we re-edit scenes by adding or removing dynamic objects, render counterfactual images, regenerate annotations, and filter edits below $0.3\%$ mask area. This yields \textbf{SceneForge-Removal (S)} with $13{,}419$ edited single-view scene instances / images from $2{,}358$ rooms and a $12{,}734/685$ train/test split, and \textbf{SceneForge-Removal (M)} with $3{,}999$ edited multi-view scene instances / images from $330$ rooms and a $3{,}803/196$ split. The latter further contains $642$ \texttt{base\_camera}, $1{,}415$ \texttt{camera\_similar}, and $1{,}942$ \texttt{camera\_diff} examples. We also derive \textbf{SceneForge-IndoorDynamicRemoval}, a scene-level dataset with $2{,}435$ room-level scenes / images in which all dynamic objects are removed, split into $2{,}313/122$ train/test examples.

\subsection{Compositional Layer Decomposition}
\label{sec:exp-layers}

\begin{figure*}[t]
    \centering
    \includegraphics[width=0.98\textwidth]{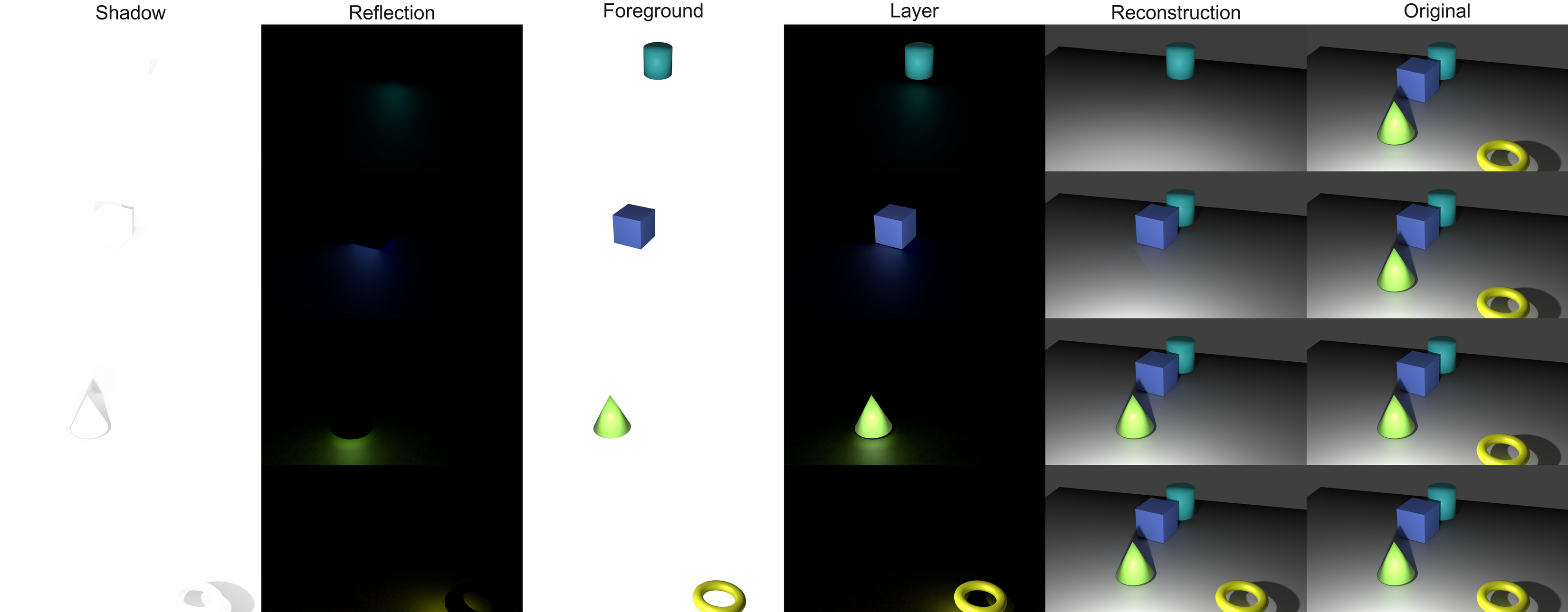}
    \caption{Visual-effect-aware layer decomposition and recomposition in Blender scenes. Rows correspond to individual dynamic objects. Columns show object-linked \textbf{shadow} layers, direct \textbf{light} contributions, object \textbf{cutouts}, projected \textbf{layers}, per-object \textbf{reconstructions}, and the final \textbf{original} render. The figure illustrates that SceneForge can decompose a scene into aligned object and effect layers, and then recompose them into the full image while taking visual effects into account.}
    \label{fig:physics-decomposition}
\end{figure*}

SceneForge produces compositional, object-linked layers rather than only edited image pairs. Consistent with Section~\ref{sec:framework}, these layers come from the same maintained world state and support instance isolation, dependency-aware grouping, amodal completion, and visual-effect-aware channels. Figure~\ref{fig:infinigen-layer-demo} shows an Infinigen indoor example: SceneForge can isolate editable objects, merge support-linked entities into shared layers, and recover occluded foreground content when hidden regions remain known in the world state. Figure~\ref{fig:physics-decomposition} complements this with a Blender scene that also decomposes cast shadows, reflected light, and other projected interactions. Because these layers come from separate Cycles passes of the same \texttt{.blend}, they can be recomposed front-to-back according to depth-map ordering to recover a render nearly identical to the original image, while preserving aligned text labels grounded in the same scene state.


\subsection{Intervention-Driven Counterfactual Generation}
\label{sec:exp-align}

Beyond decomposition quality, SceneForge is intended to produce counterfactual supervision from world-state interventions rather than independent image edits. We therefore analyze the re-edited results that form \textbf{SceneForge-Removal}, focusing on counterfactual object-removal results in both single-camera settings \textbf{SceneForge-Removal (S)} and registered multi-camera observations of the same edited world state \textbf{SceneForge-Removal (M)}.

\begin{figure}[t]
    \centering
    \includegraphics[width=0.98\linewidth]{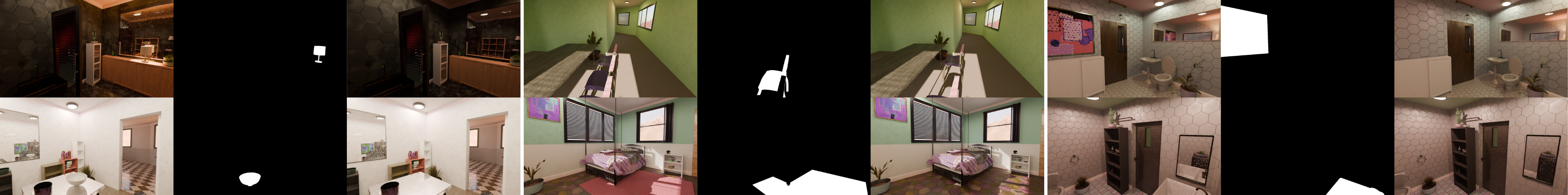}
    \caption{Single-camera counterfactual examples from \textbf{SceneForge-Removal (S)}. Columns show the original image, removal mask, and counterfactual render after object removal. SceneForge produces coherent counterfactual supervision beyond the masked region, including revealed background content and consistent scene-level effects such as shadows and illumination.}
    \label{fig:counterfactual-triplets}
\end{figure}

\begin{figure*}[t]
    \centering
    \includegraphics[width=\textwidth]{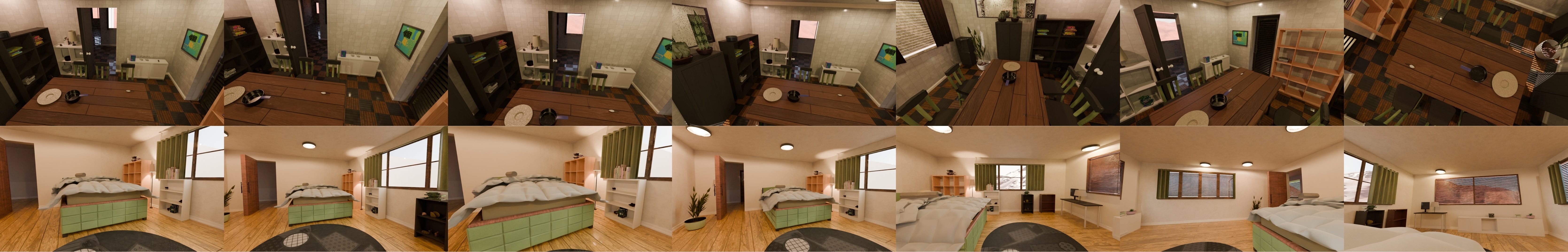}
    \caption{Registered multi-camera counterfactual strips from \textbf{SceneForge-Removal (M)}. Each strip shows a scene rendered from a base camera (left) and six registered viewpoints from the same edited world state. Columns 2--4 show nearby high-overlap views, while Columns 5--7 show more diverse viewpoints. Revealed background regions and visual effects such as shadows and reflections remain spatially consistent across views as all renders are derived from the same persistent 3D world state.}
    \label{fig:multiview-counterfactual-strips}
\end{figure*}

\textbf{Qualitative counterfactual evidence.} We examine thin structures, specular surfaces, and soft penumbrae to assess whether SceneForge re-edits remain consistent after object removal. Figure~\ref{fig:counterfactual-triplets} shows single-camera counterfactual examples from \textbf{SceneForge-Removal (S)}. By deriving targets from 3D state transitions rather than image-space heuristics, SceneForge preserves revealed background and updated illumination. Figure~\ref{fig:multiview-counterfactual-strips} shows registered multi-camera strips from \textbf{SceneForge-Removal (M)}. To provide spatial consistency, we sample viewpoints around the base camera with a tiered protocol: one group provides high-overlap views ($>70\%$) to simulate local continuity, while the other provides more diverse perspectives (e.g., bird's-eye and corner views). Candidate cameras are filtered using same-room Line-of-Sight checks, wall-clearance constraints, and 7-point bounding-box visibility verification. Because these views render a shared updated world state, all counterfactual signals are aligned by construction. The formulation also extends to synchronized counterfactual video across registered viewpoints, as shown in the appendix.

\subsection{Downstream Applications: Object-Level and Scene-Level Removal}
\label{sec:exp-downstream}

Finally, we study two downstream removal tasks under matched settings: object-level removal and scene-level removal. We use \textbf{Qwen-Image-Edit-2509} as the editing backbone and fine-tune the same pretrained checkpoint with LoRA on mask-conditioned \texttt{orig/mask/target} triplets. Across the three object-removal settings, we keep architecture, optimization, compute, and training duration fixed, so the only changing factor is the training supervision itself. All reported training runs use $8\times$ NVIDIA H200 GPUs and take about $20$ hours. As public supervision, we use a lightweight mixture of existing object-removal resources, including RORem \citep{li2025rorem}, OmniEraser \citep{wei2025omnieraser}, and MuLAn \citep{tudosiu2024mulan}. Additional training details are provided in the supplementary material.

\textbf{Case study 1: Object-level removal.} We compare three settings: \textbf{OpenData-30K}, which uses $30$K public training samples; \textbf{OpenData+SceneForge-30K}, which keeps the same overall budget but replaces part of the public data with SceneForge samples; and \textbf{SceneForge-16K}, which uses only SceneForge training data. Because all three runs share the same pretrained model and fine-tuning recipe, the comparison isolates the effect of supervision quality. We evaluate on four paired test sets: \textbf{SceneForge-Removal (S) test} ($685$ images), \textbf{SceneForge-Removal (M) test} ($196$ images), \textbf{RORem test\_300} ($300$ images), and \textbf{RemovalBench} ($69$ images). Because the publicly released RORem test set does not provide ground-truth targets, we randomly select $300$ paired examples from RORem as an additional test subset for evaluation. We report full-image \textbf{PSNR}, \textbf{SSIM}, and \textbf{LPIPS} to measure not only object removal within the mask but also restoration of correlated visual effects such as shadows and newly revealed background.

\begin{table*}[t]
  \caption{Object-removal results of three settings with or without SceneForge.}
  \label{tab:downstream-removal-main}
  \begin{center}
    \small
    \resizebox{\textwidth}{!}{%
    \begin{tabular}{l|lccc}
      \toprule
      Test set & Metric & OpenData-30K & OpenData+SceneForge-30K & SceneForge-16K \\
      \midrule
      \multirow{3}{*}{SceneForge-Removal (S) test} & PSNR $\uparrow$ & 33.4079 & 37.2342 & \textbf{38.3180} \\
       & SSIM $\uparrow$ & 0.9602 & 0.9795 & \textbf{0.9824} \\
       & LPIPS $\downarrow$ & 0.0360 & 0.0201 & \textbf{0.0157} \\
      \midrule
      \multirow{3}{*}{SceneForge-Removal (M) test} & PSNR $\uparrow$ & 30.4853 & 33.4978 & \textbf{34.7007} \\
       & SSIM $\uparrow$ & 0.9469 & 0.9641 & \textbf{0.9699} \\
       & LPIPS $\downarrow$ & 0.0529 & 0.0352 & \textbf{0.0272} \\
      \midrule
      \multirow{3}{*}{RORem test\_300} & PSNR $\uparrow$ & 19.4663 & 19.5509 & \textbf{20.3384} \\
       & SSIM $\uparrow$ & 0.8508 & 0.8557 & \textbf{0.8803} \\
       & LPIPS $\downarrow$ & 0.1622 & 0.1538 & \textbf{0.1328} \\
      \midrule
      \multirow{3}{*}{RemovalBench} & PSNR $\uparrow$ & 26.3099 & 27.3893 & \textbf{28.7500} \\
       & SSIM $\uparrow$ & 0.8457 & 0.8482 & \textbf{0.8584} \\
       & LPIPS $\downarrow$ & 0.1242 & 0.1144 & \textbf{0.0870} \\
      \bottomrule
    \end{tabular}}
  \end{center}
\end{table*}

Table~\ref{tab:downstream-removal-main} shows a clear trend. \textbf{OpenData+SceneForge-30K} improves substantially over \textbf{OpenData-30K}, indicating that SceneForge supervision brings consistent gains for object-level removal across both in-domain and external benchmarks. Moreover, \textbf{SceneForge-16K} achieves the best result on every reported metric for all four evaluation sets despite using fewer total samples than either $30$K setting. This suggests that cleaner, more intervention-consistent supervision can outweigh larger but more mixed training pools.

\begin{figure*}[t]
    \centering
    \makebox[\textwidth][c]{%
        \includegraphics[width=0.5\textwidth]{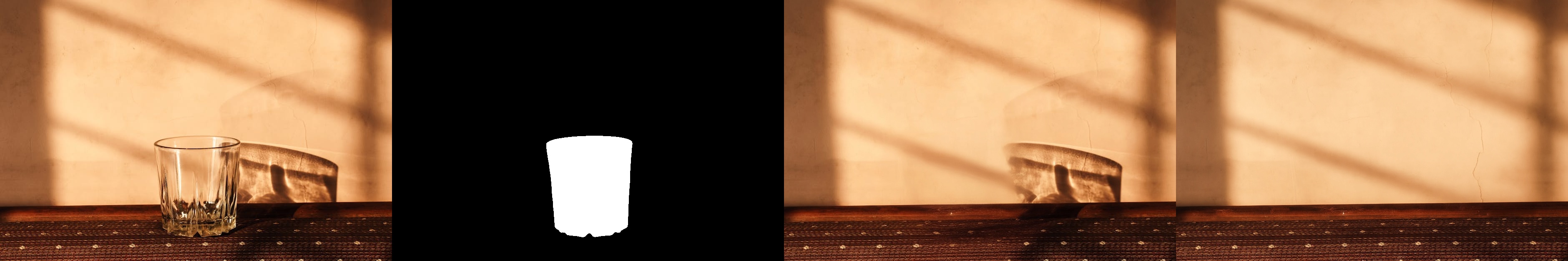}%
        \includegraphics[width=0.5\textwidth]{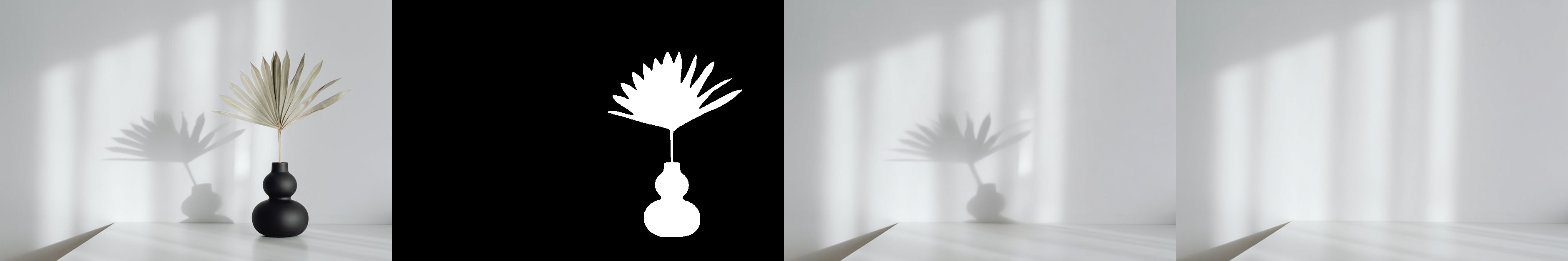}}
    \par\nointerlineskip
    \makebox[\textwidth][c]{%
        \includegraphics[width=0.5\textwidth]{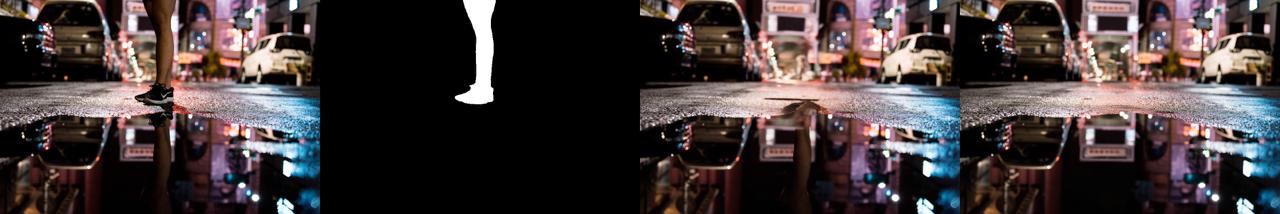}%
        \includegraphics[width=0.5\textwidth]{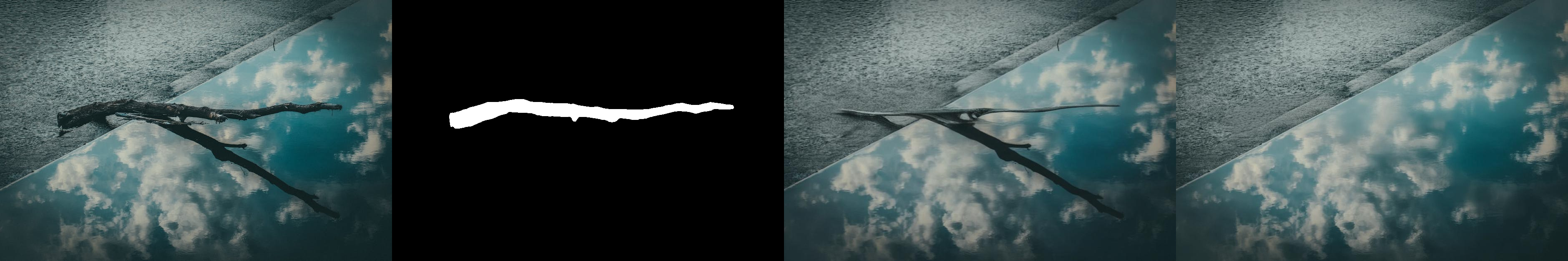}}
    \par\nointerlineskip
    \makebox[\textwidth][c]{%
        \includegraphics[width=0.5\textwidth]{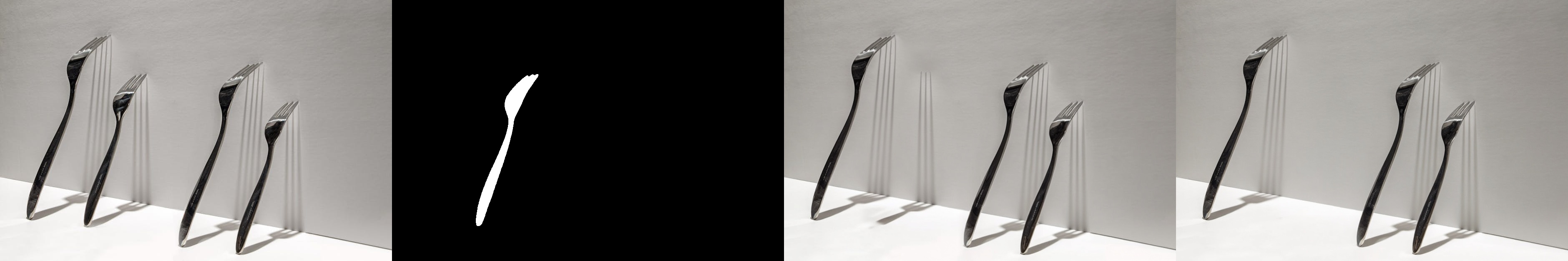}%
        \includegraphics[width=0.5\textwidth]{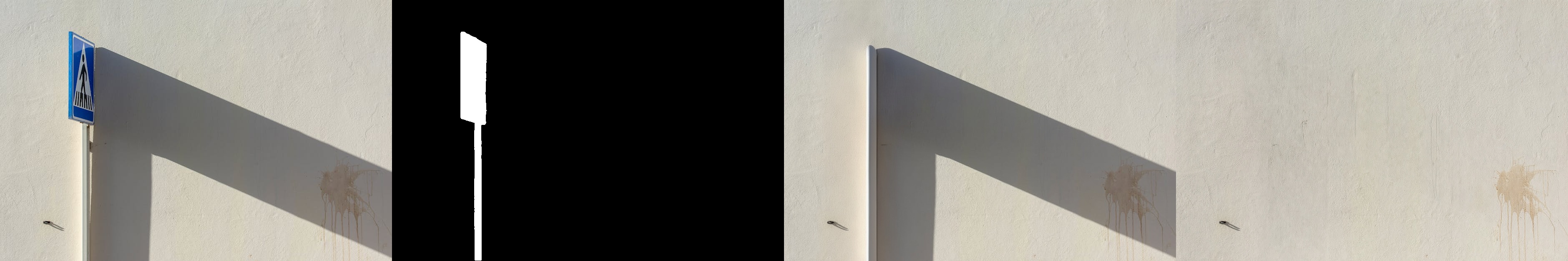}}
    \caption{Hard-case examples from \textbf{Removal-HardEffects}. Within each example, columns from left to right show the original image, mask, \textbf{OpenData-30K} result, and \textbf{SceneForge-16K} result. These qualitative comparisons show that \textbf{SceneForge-16K} more reliably removes associated shadows, reflections, and other local side effects during object removal.}
    \label{fig:removal-hardeffects-userstudy}
\end{figure*}

We curate \textbf{Removal-HardEffects}, a set of $60$ open-source images~\cite{pexels2025} with strong shadows or reflections and manual masks, as a qualitative benchmark for hard real-world cases. Figure~\ref{fig:removal-hardeffects-userstudy} shows comparisons where SceneForge-trained models more reliably remove associated shadows, reflections, and other local side effects.

\begin{figure*}
    \centering
    \includegraphics[width=\textwidth]{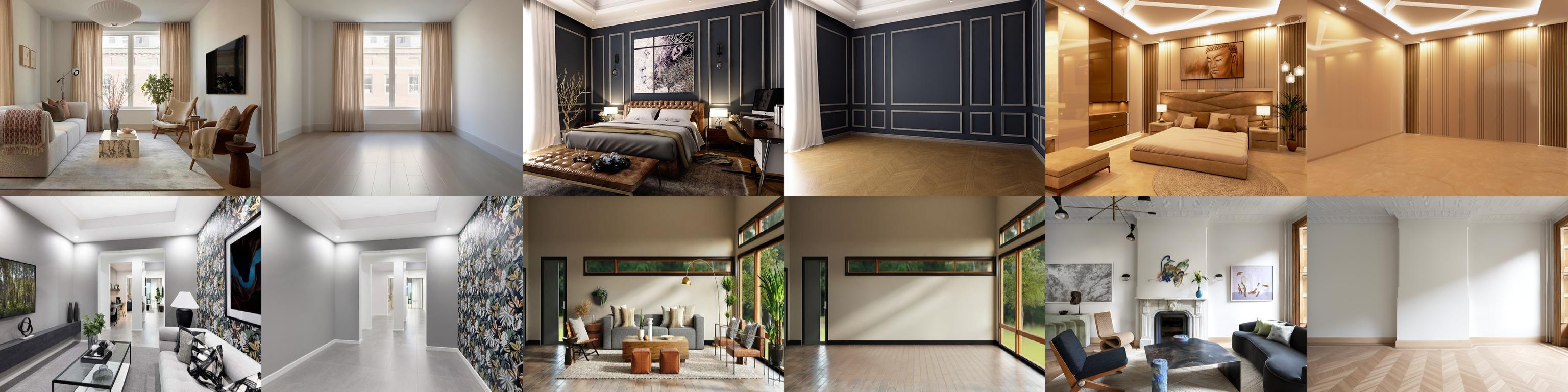}
    \caption{Subjective results on realistic real-world cases using the model trained on \textbf{SceneForge-IndoorDynamicRemoval}. Six examples are arranged in two rows with three resized, uncropped cases per row.}
    \label{fig:downstream-structure-qualitative}
\end{figure*}

\textbf{Case study 2: Scene-level removal.} We also study scene-level removal, where all dynamic indoor objects are removed while preserving architectural structure. Training the same LoRA backbone on \textbf{SceneForge-IndoorDynamicRemoval} ($2{,}313/122$ train/test) yields full-image scores of \textbf{30.594} PSNR, \textbf{0.9498} SSIM, and \textbf{0.0477} LPIPS. Figure~\ref{fig:downstream-structure-qualitative} shows qualitative results on real-world cases.

\section{Discussion and Limitations}

SceneForge offers a structured route to multimodal supervision when consistency under intervention matters, since one edited world state can yield aligned counterfactual observations, decomposition channels, annotations, and multi-view supervision. Its usefulness still depends on world-state fidelity, dependency propagation, rendering quality, and asset licensing, and synthetic supervision does not remove domain shift. Because our empirical study focuses on removal tasks, we view SceneForge as a flexible supervision engine that complements rather than replaces real or hybrid datasets.

\section{Conclusion}

We presented SceneForge, a framework for structured world supervision from 3D interventions. By modeling scenes as editable world states with explicit semantic, geometric, and physical dependencies, SceneForge generates aligned counterfactual images, decomposition channels, annotations, relations, and geometry-aware signals from state transitions rather than observation-level heuristics. Our results suggest that this world-centric formulation provides more consistent supervision and stronger downstream utility for multimodal learning, and we hope it helps establish editable 3D worlds as a practical foundation for future multimodal data engines.

\newpage
\bibliographystyle{plainnat}
\bibliography{refs}

\clearpage
\appendix
\section*{Supplementary}

\subsection*{SceneForge engine implementation details}

The released SceneForge engine provides a runnable path from editable scenes to released supervision. In our current instantiation, it starts from Infinigen indoor scenes and editable \texttt{.blend} assets, then uses Blender Cycles multi-pass rendering to produce full-scene renders, structure-only renders, object-removal counterfactuals, and object-linked decomposition outputs from the same underlying scene asset. In this way, the released implementation concretely realizes the pipeline described in Section~\ref{sec:framework} without requiring a separate symbolic world-state database at inference time.

A practical implementation detail is that the dynamic-versus-structure distinction is operationalized at render time. The engine uses scene metadata and object categories to decide which entities are treated as editable dynamic objects and which are kept as structural scene elements, then applies the corresponding pass configuration when constructing object-removal or structural-only supervision. This allows the same editable scene to support both object-level and scene-level counterfactual generation under a single rendering pipeline.

For multi-view counterfactual construction, candidate cameras are not sampled uniformly and used directly. Instead, SceneForge first proposes nearby and diverse viewpoints around a base camera, then filters them with same-room Line-of-Sight checks, wall-clearance constraints, and 7-point target-visibility verification. In the released multi-view corpus, this yields $2{,}772$ registered multi-view scene instances under $8$ camera settings in total. These constraints keep the retained views tied to the same edited world state while avoiding degenerate cameras that either lose the edited object entirely or violate simple scene geometry constraints.

For object-removal data, interventions are also filtered after rendering. We discard tiny edits whose valid removal region is below the mask-area threshold, and we keep only examples whose edited target remains consistent with the propagated scene state. Post-processing then exports released \texttt{orig/mask/target} triplets together with mask-coverage metadata. For layer supervision, SceneForge stores object-linked decomposition outputs from the same maintained scene state, including per-object components, dependency-aware groupings, effect-aware layers, and amodal targets when occluded structure can be recovered from the editable world representation. This shared-state construction is what allows a single intervention to produce aligned RGB targets, decomposition signals, and cross-view supervision without post hoc correspondence matching.

Optional LLM-assisted scene querying used in some internal builds is therefore not required to run the released code path. In the released implementation, the core multi-view and visibility checks are handled geometrically inside the engine itself.

\subsection*{Statistics provenance and verification}

Table~\ref{tab:dataset_stats} mixes two kinds of statistics. The first two rows, namely \textbf{SceneForge (S, raw layers)} and \textbf{SceneForge (M, raw layers)}, are reported from the full generation manifests used for dataset construction. In our release tree, these raw-corpus counts correspond to the manifest summary in \texttt{data/paper\_stats.json} together with the updated multi-view corpus summary, which gives $25{,}688$ layers / $2{,}369$ rooms / single-view scene instances for \textbf{SceneForge (S)} and $34{,}865$ image layers / $517$ rooms / $2{,}772$ registered multi-view scene instances for \textbf{SceneForge (M)}, with the latter spanning $8$ camera settings.

By contrast, the filtered downstream datasets in Table~\ref{tab:dataset_stats} and Section~\ref{sec:dataset-stats} are directly verifiable from the released JSONL split files. In particular, \textbf{SceneForge-Removal (S)} contains $12{,}734$ train and $685$ test examples, totaling $13{,}419$ edited single-view scene instances from $2{,}358$ rooms; \textbf{SceneForge-Removal (M)} contains $3{,}803$ train and $196$ test examples, totaling $3{,}999$ edited multi-view scene instances from $330$ rooms; and \textbf{SceneForge-IndoorDynamicRemoval} contains $2{,}313$ train and $122$ test examples, totaling $2{,}435$ room-level scenes. The single-camera and multi-camera room sets remain disjoint under a UUID-based room key.

A practical source of confusion is that the release tree may also contain export-side summaries such as \texttt{batch\_export\_summary.json}. These files summarize export jobs at the scene level and are useful for sanity checking, but they are not one-row-per-layer manifests, so their \texttt{ok} or \texttt{object\_folders} aggregates are not expected to match the raw-layer totals in Table~\ref{tab:dataset_stats} exactly. For this reason, we use the full generation manifests for the raw-corpus rows and the released JSONL splits for the filtered downstream rows.

\subsection*{Additional downstream training details}

For object-level removal, all three settings in Section~\ref{sec:exp-downstream} start from the same \textbf{Qwen-Image-Edit-2509} pretrained checkpoint and use the same mask-conditioned \texttt{orig/mask/target} training objective with LoRA fine-tuning. We keep architecture, resolution, batch size, optimizer / scheduler, GPU count, and wall-clock training budget aligned across settings, and we do not introduce dataset-specific architectural changes, extra pretraining, or benchmark-specific retuning. The comparison is therefore intended to isolate the effect of the training supervision itself. \textbf{OpenData-30K} contains exactly $30{,}000$ public triplets: \textbf{RORem Final-HR} ($2{,}463$), \textbf{Final\_open\_RORem} ($4{,}500$), the public \textbf{OmniEraser} subset ($5{,}008$), \textbf{MuLAn} \texttt{coco\_init} ($9{,}000$), and \textbf{MuLAn} \texttt{laion\_init} ($9{,}029$).

\textbf{OpenData+SceneForge-30K} keeps the same overall budget of $30{,}000$ triplets, retains \textbf{RORem Final-HR} in full, and replaces part of the public supervision with all currently used SceneForge object-removal triplets. Concretely, this setting contains \textbf{RORem Final-HR} ($2{,}463$), \textbf{SceneForge-Removal (S) train} ($12{,}734$), \textbf{SceneForge-Removal (M) train} ($3{,}803$), \textbf{Final\_open\_RORem} ($1{,}798$), the public \textbf{OmniEraser} subset ($2{,}000$), \textbf{MuLAn} \texttt{coco\_init} ($3{,}595$), and \textbf{MuLAn} \texttt{laion\_init} ($3{,}607$), again totaling $30{,}000$ triplets. This matched-budget design is meant to test whether replacing part of the public pool with SceneForge improves supervision quality, rather than simply increasing the amount of training data.

\textbf{SceneForge-16K} uses only SceneForge triplets for object-level removal. In our current setup, this corresponds to the union of \textbf{SceneForge-Removal (S) train} and \textbf{SceneForge-Removal (M) train}, for $16{,}537$ triplets in total; we refer to it as \textbf{SceneForge-16K} for brevity. For scene-level removal, we use the same backbone and optimization recipe, but train only on \textbf{SceneForge-IndoorDynamicRemoval}, with $2{,}313$ training images and $122$ test images.

\subsection*{Additional layer decomposition examples}

\begin{figure*}[t]
    \centering
    \includegraphics[width=\textwidth]{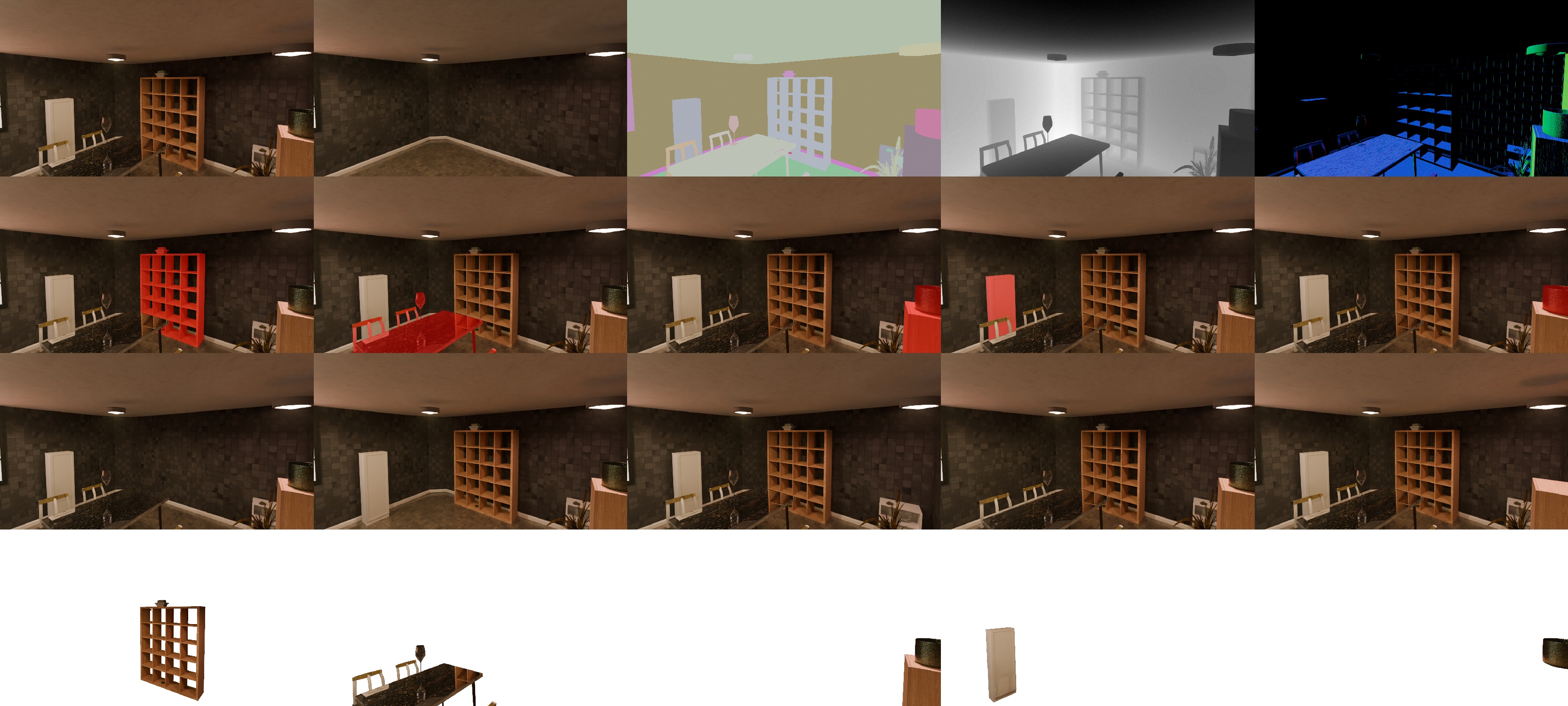}
    \par\smallskip
    \includegraphics[width=\textwidth]{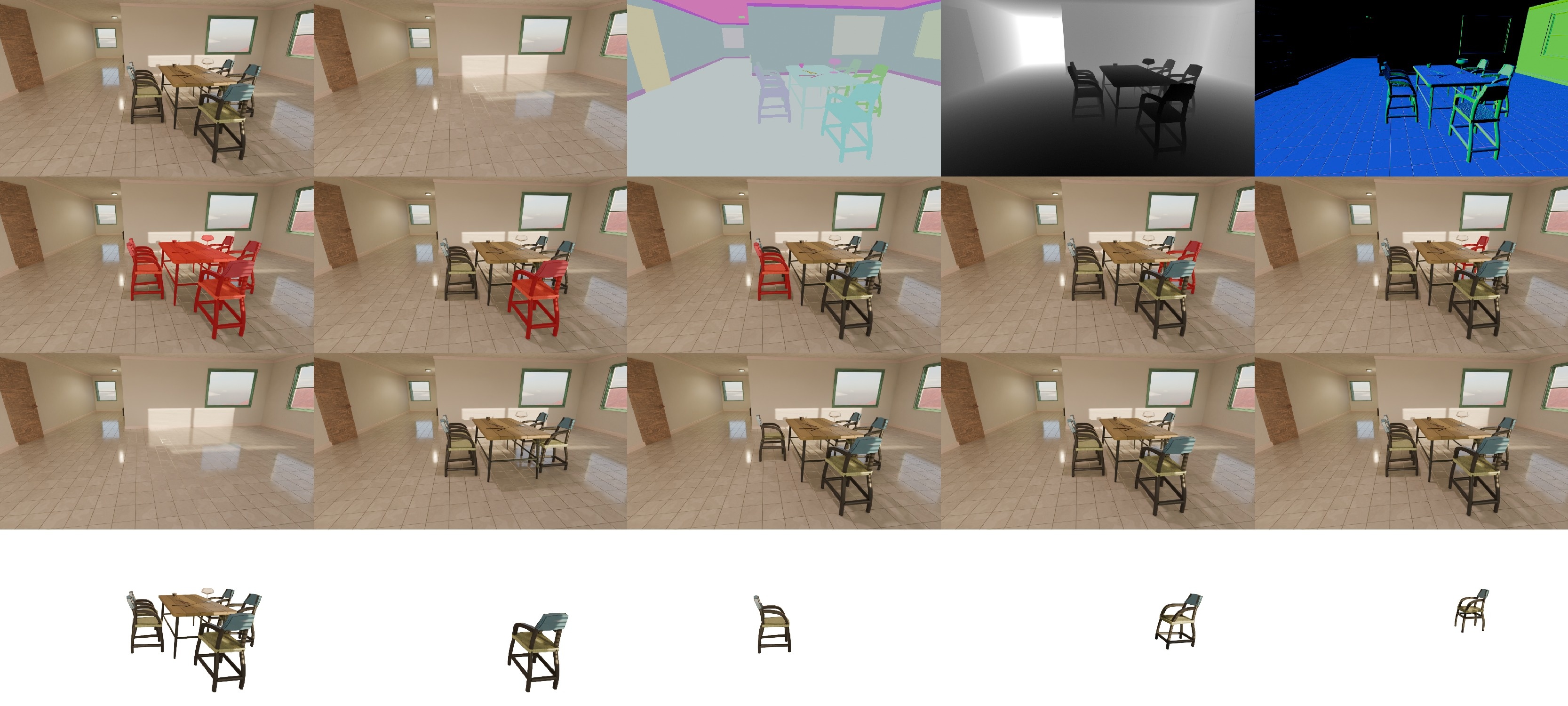}
    \par\smallskip
    \includegraphics[width=\textwidth]{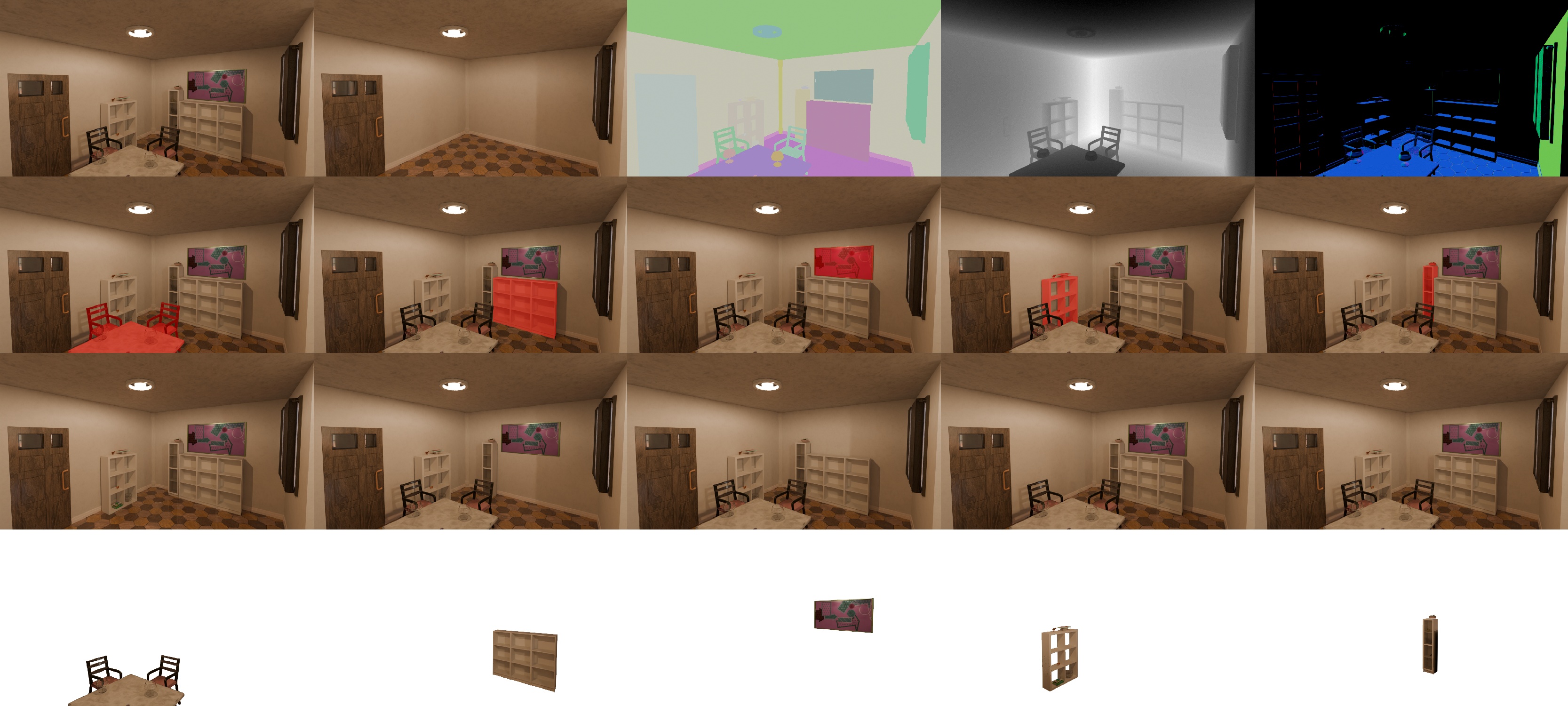}
    \caption{Three additional layer decomposition examples. Each example is shown without cropping and illustrates scene-aligned decomposition and intervention-conditioned outputs derived from the same maintained world representation. Only five representative layers are shown due to page limits.}
    \label{fig:infinigen-layer-demo-supp-a}
\end{figure*}

\begin{figure*}[t]
    \centering
    \includegraphics[width=\textwidth]{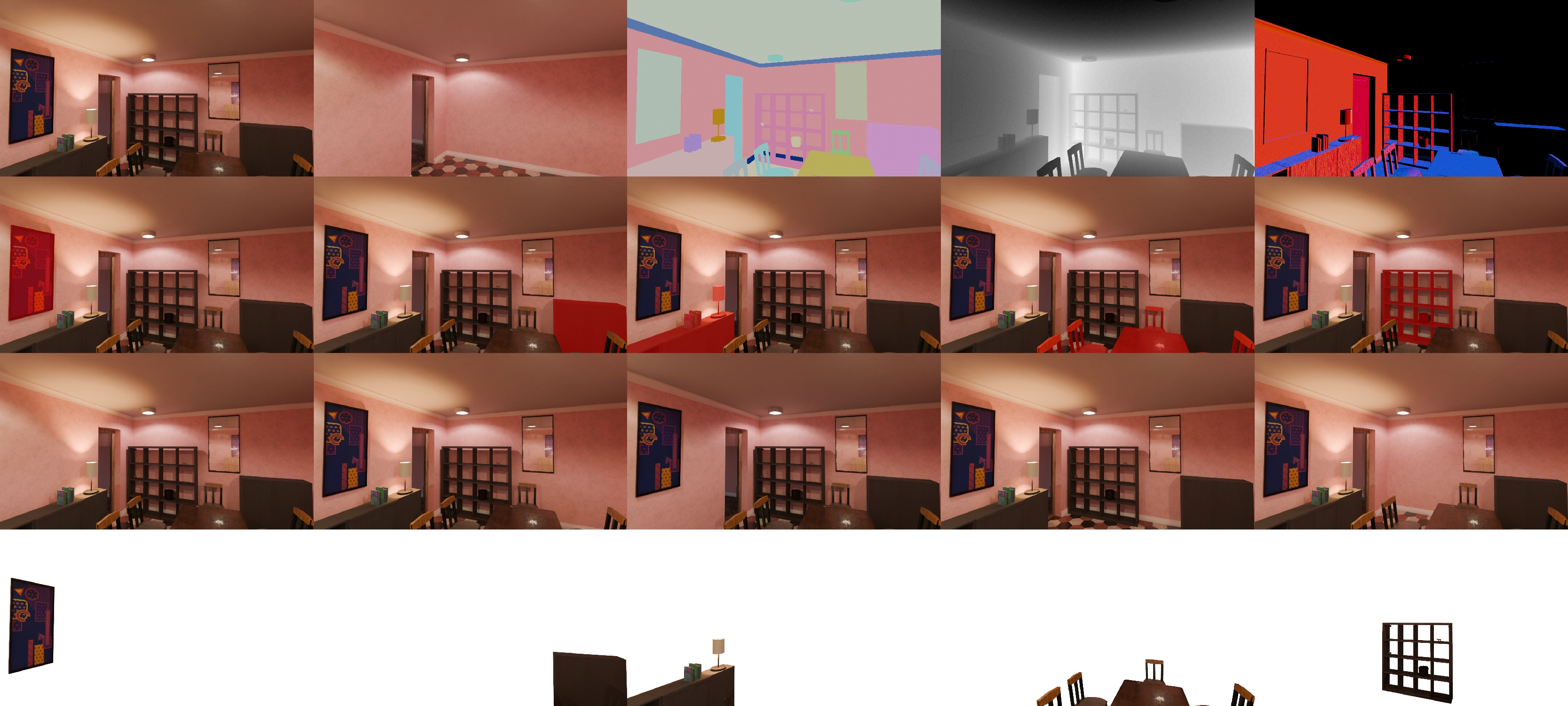}
    \par\smallskip
    \includegraphics[width=\textwidth]{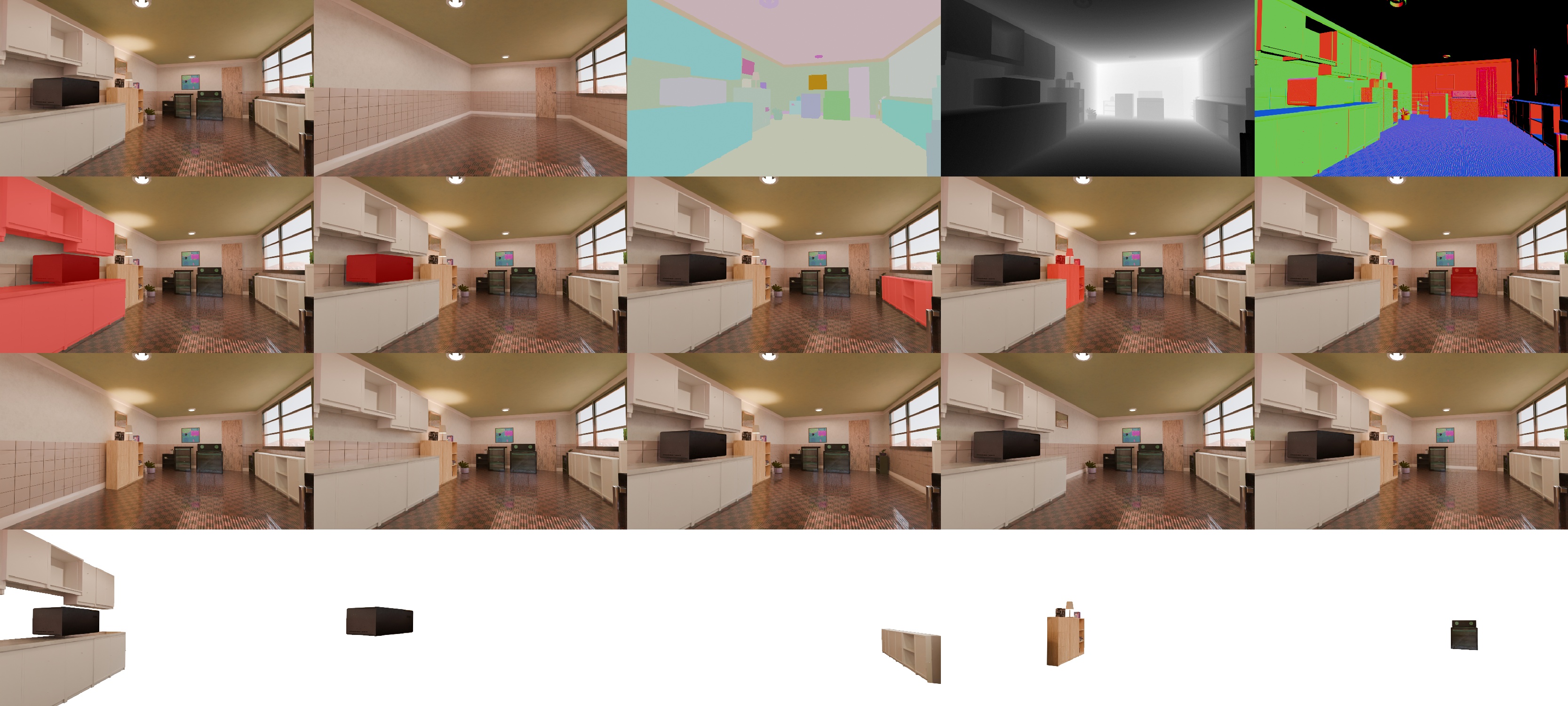}
    \par\smallskip
    \includegraphics[width=\textwidth]{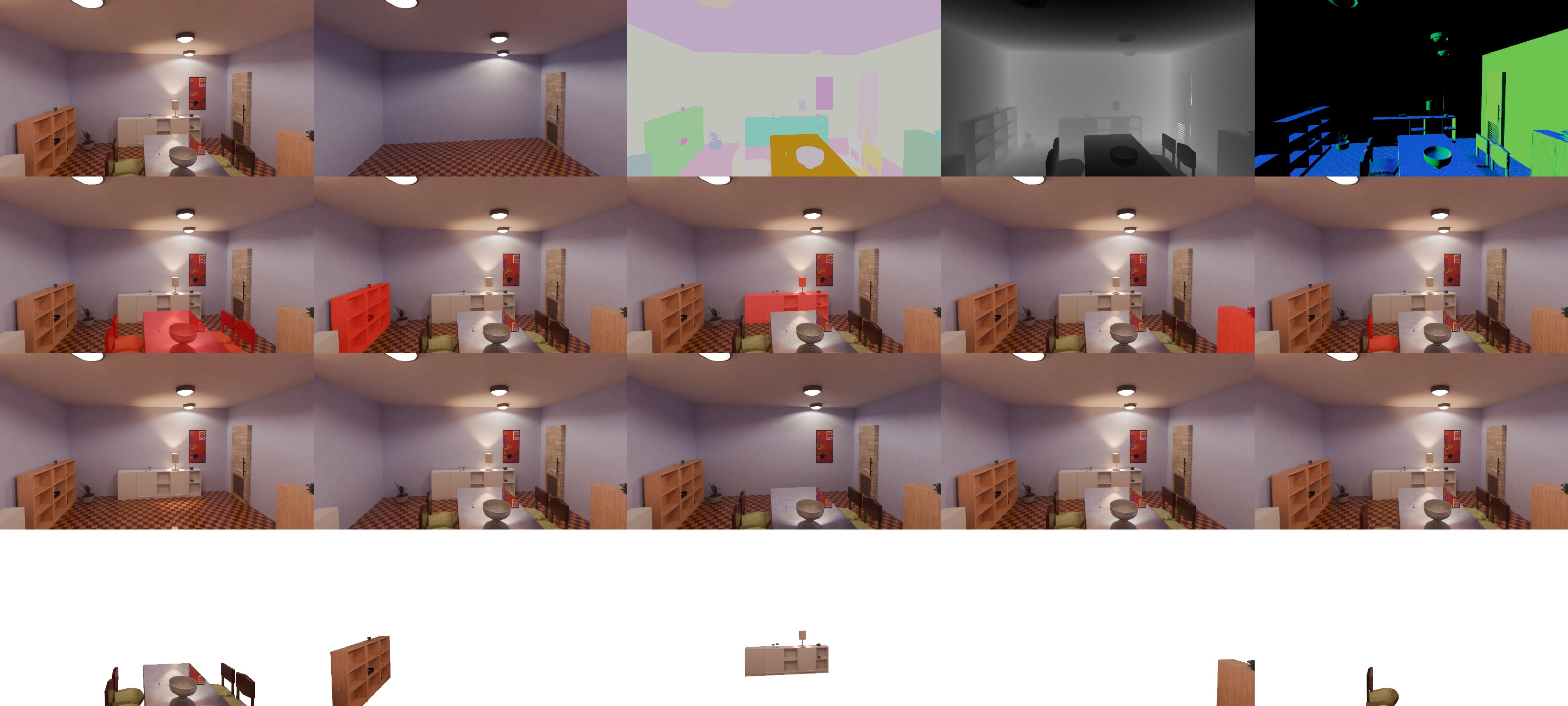}
    \caption{Three further layer decomposition examples. These cases further show that the same maintained world state can support aligned structure-aware signals and object-level decomposition outputs across different room configurations. Only five representative layers are shown due to page limits.}
    \label{fig:infinigen-layer-demo-supp-b}
\end{figure*}

Figures~\ref{fig:infinigen-layer-demo-supp-a} and~\ref{fig:infinigen-layer-demo-supp-b} provide additional layer decomposition examples complementary to Figure~\ref{fig:infinigen-layer-demo} in the main paper. Across different indoor layouts, these examples illustrate that object-linked decomposition remains available together with scene-aligned structural signals, intervention-conditioned renders, and isolated object components derived from the same maintained world state.

\subsection*{Additional visual-effect-aware decomposition examples}

These additional qualitative examples complement Figure~\ref{fig:physics-decomposition} in the main paper and further illustrate SceneForge's visual-effect-aware decomposition and reconstruction behavior across diverse indoor scenes.

\begin{figure*}[t]
    \centering
    \includegraphics[width=0.94\textwidth]{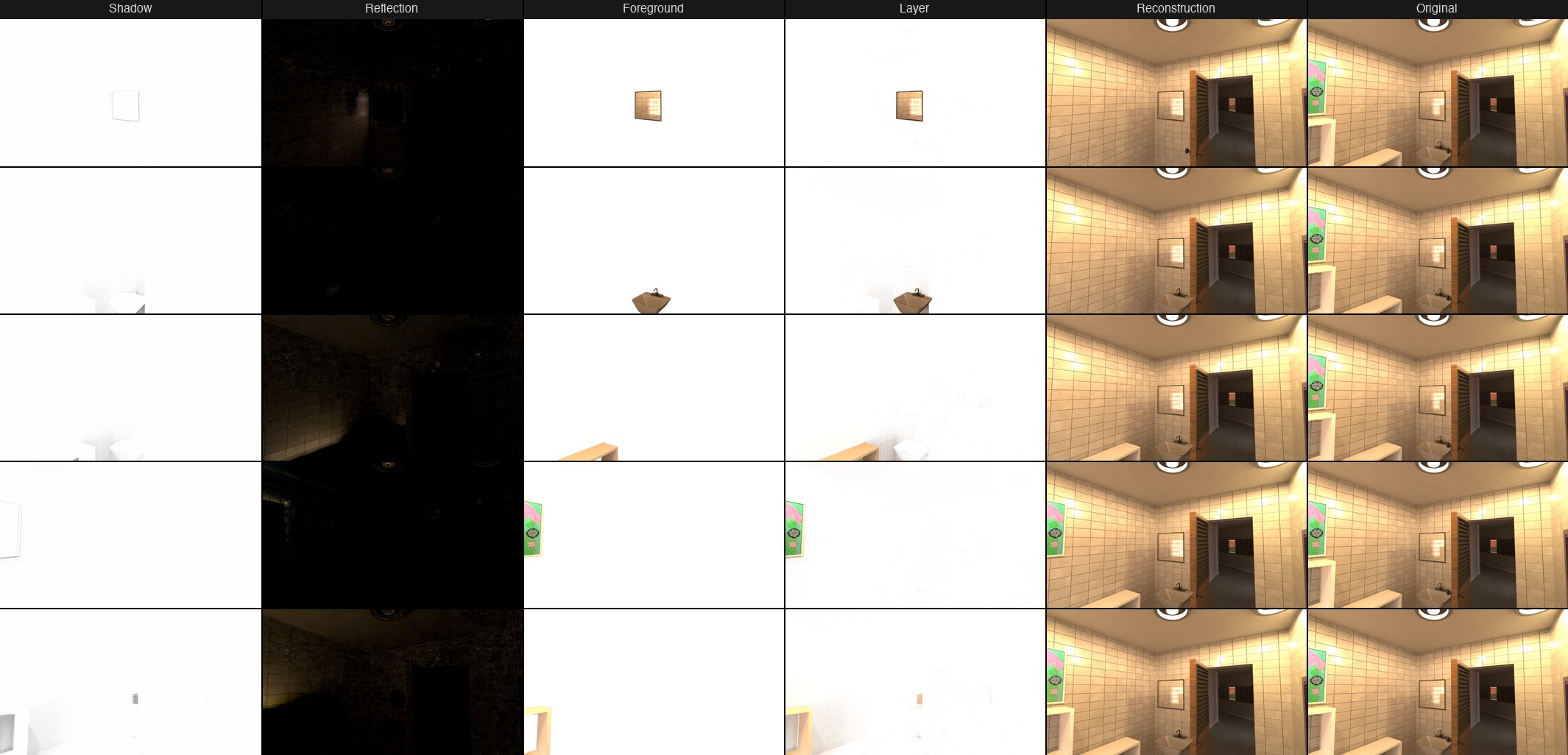}
    \par\vspace{0.2em}
    \includegraphics[width=0.94\textwidth]{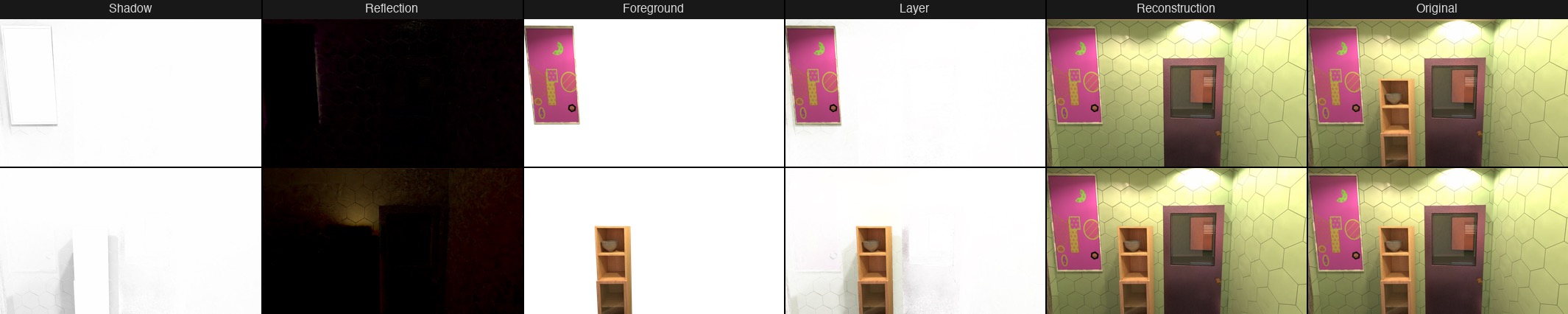}
    \par\vspace{0.2em}
    \includegraphics[width=0.94\textwidth]{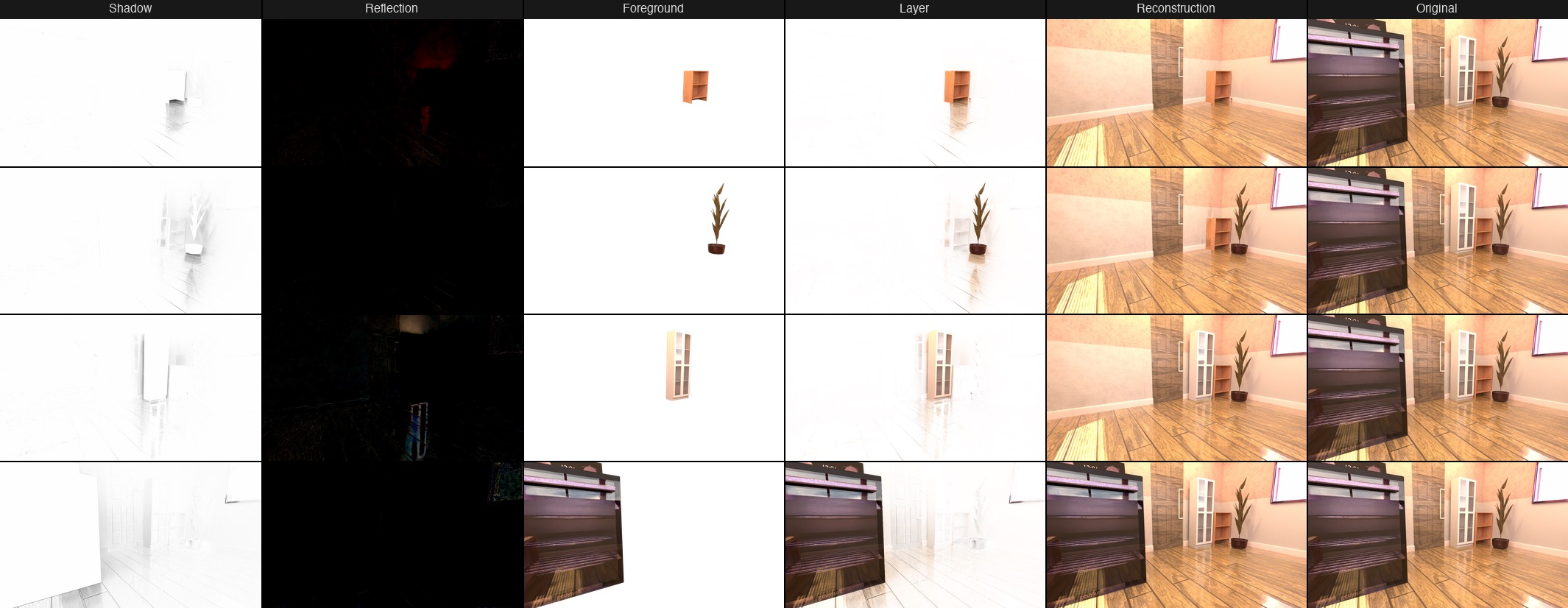}
    \par\vspace{0.2em}
    \includegraphics[width=0.94\textwidth]{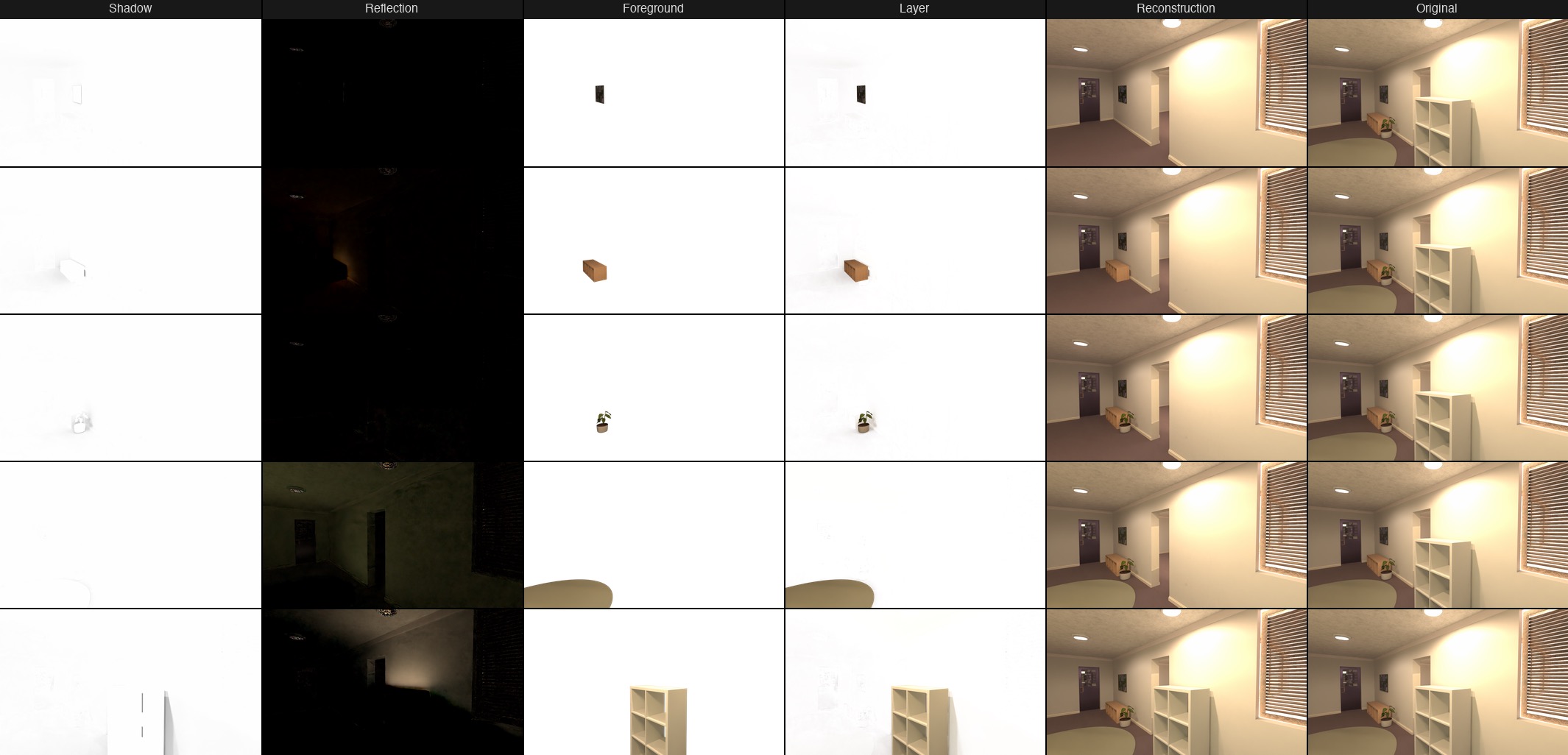}
    \caption{Additional visual-effect-aware decomposition examples complementary to Figure~\ref{fig:physics-decomposition}. Columns show the separated shadow, reflection, foreground, and layer components, together with the resulting reconstruction and original image.}
    \label{fig:visual-inf-supp}
\end{figure*}

\subsection*{Single-camera counterfactual examples}

\begin{figure}[t]
    \centering
    \includegraphics[width=\linewidth]{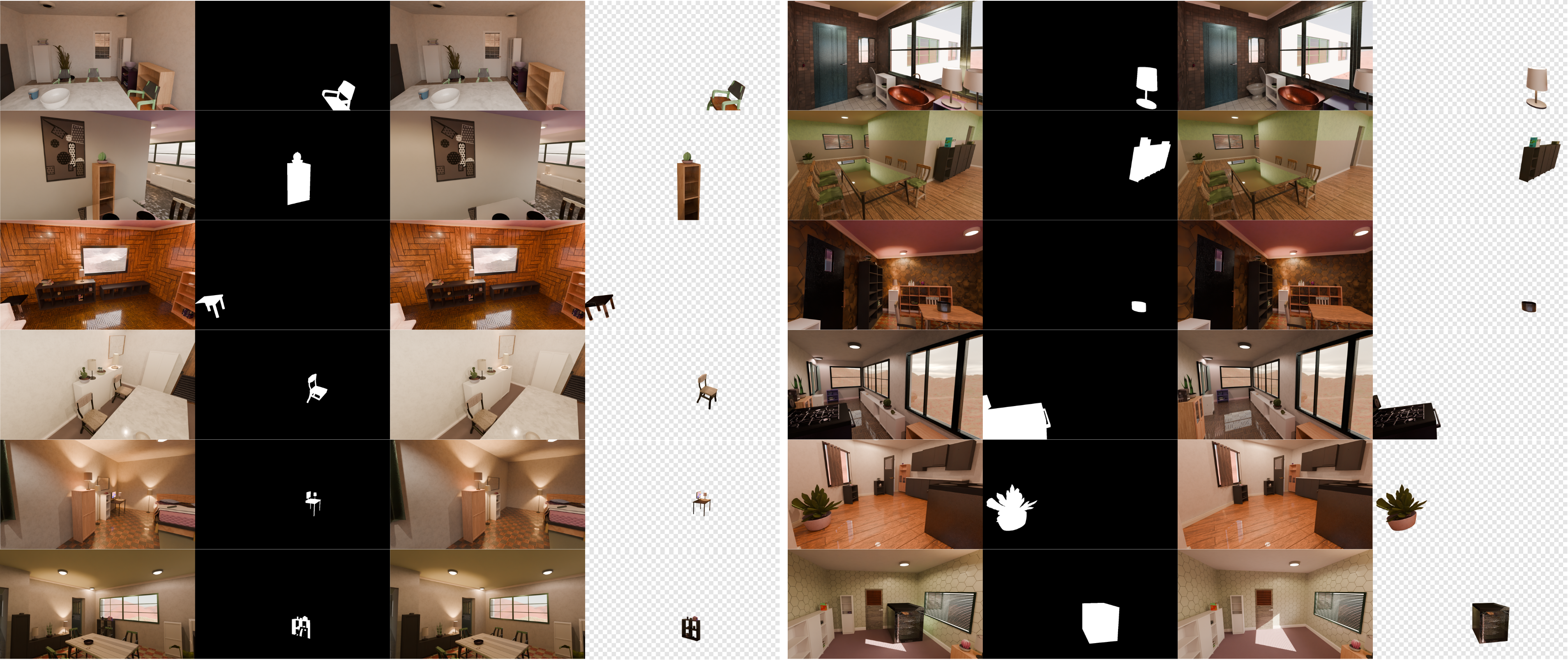}
    \caption{Additional counterfactual examples from \textbf{SceneForge-Removal (S)}. Columns show the original image, removal mask, counterfactual image after removal, and aligned RGBA amodal layers of the removed object derived from the same world state. For partially occluded objects (left), the amodal layers recover occluded object regions, while for fully visible objects (right), the RGBA layers coincide with the visible object appearance.}
    \label{fig:counterfactual-triplets-supp}
\end{figure}

The main paper presents a compact visualization of single-camera counterfactual supervision in Figure~\ref{fig:counterfactual-triplets}. Figure~\ref{fig:counterfactual-triplets-supp} provides additional examples together with the amodal RGBA ground truth of the removed object, showing that counterfactual targets, removal masks, and amodal object layers are generated from the same world-state intervention.


\subsection*{Synchronized Counterfactual Video Observations}

\begin{figure}[t]
    \centering
    \includegraphics[width=\linewidth]{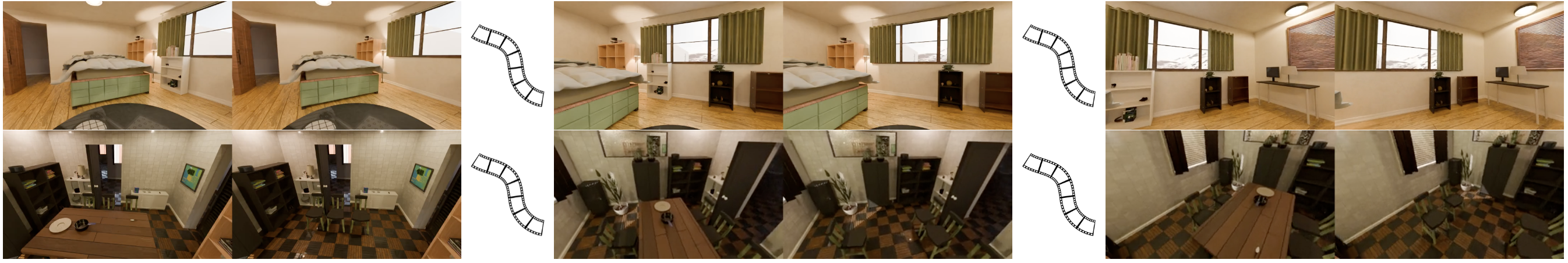}
    \caption{Examples of synchronized counterfactual video observations generated from the same edited world state. Each row shows temporally aligned frames from two registered viewpoints observing the same intervention over time. Revealed background regions and scene-level appearance changes remain consistent across viewpoints and frames because all renders are derived from a shared persistent 3D world state.}
    \label{fig:counterfactual-video-pairs}
\end{figure}

Beyond static counterfactual images, the same world-state formulation also supports synchronized counterfactual video observations across registered viewpoints. Figure~\ref{fig:counterfactual-video-pairs} shows representative examples rendered from two cameras observing the same edited world state over time. Each row corresponds to one scene, while the left and right groups show temporally synchronized frames from two registered viewpoints.

Because both videos are rendered from the same intervention and persistent world state, scene geometry, revealed background regions, and removal-induced appearance changes remain consistent across viewpoints and frames. These examples illustrate that SceneForge naturally extends from single-frame counterfactual supervision to synchronized video supervision without requiring post hoc correspondence matching.

\subsection*{Additional visual results for Section~4.4}

\begin{figure*}[t]
    \centering
    \includegraphics[width=0.92\textwidth]{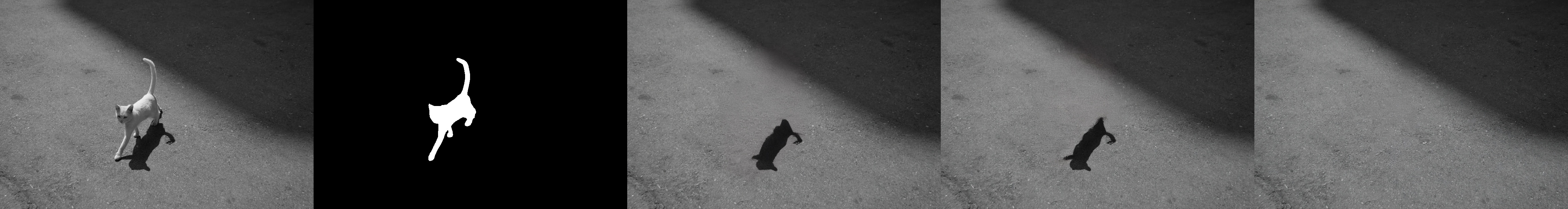}
    \par\vspace{0.05em}
    \includegraphics[width=0.92\textwidth]{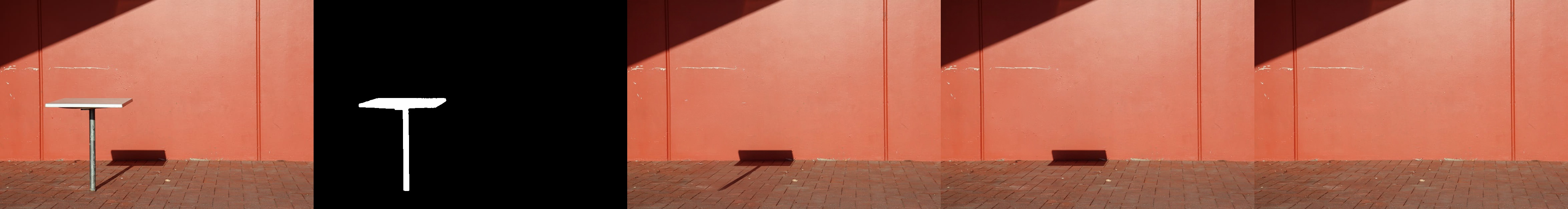}
    \par\vspace{0.05em}
    \includegraphics[width=0.92\textwidth]{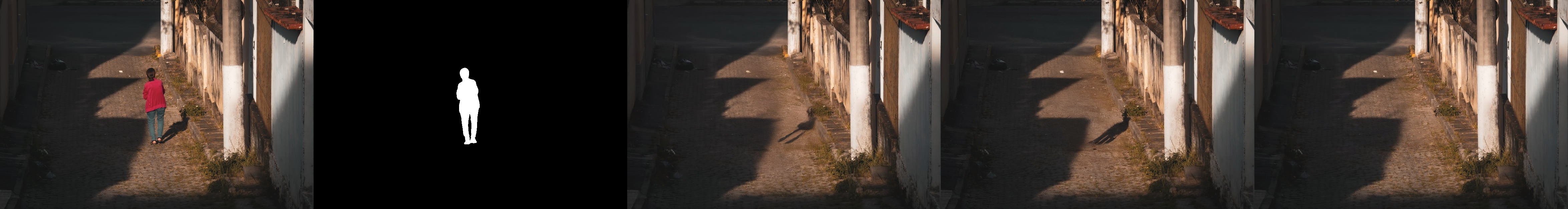}
    \par\vspace{0.05em}
    \includegraphics[width=0.92\textwidth]{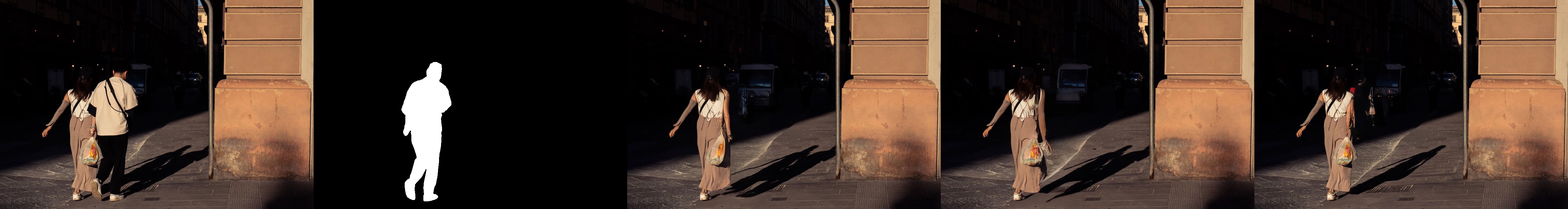}
    \par\vspace{0.05em}
    \includegraphics[width=0.92\textwidth]{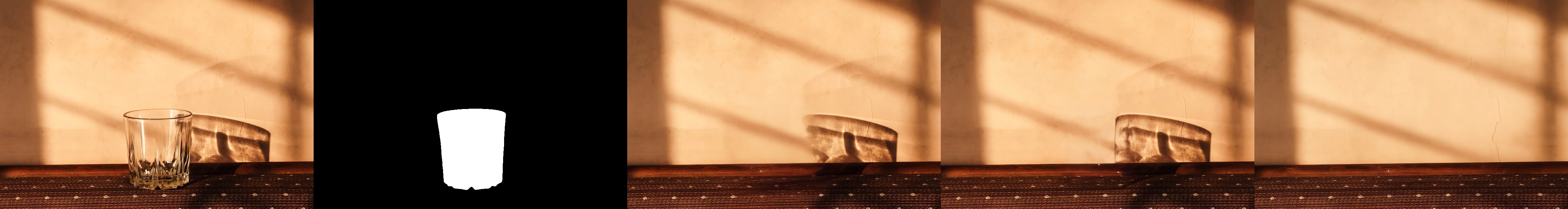}
    \par\vspace{0.05em}
    \includegraphics[width=0.92\textwidth]{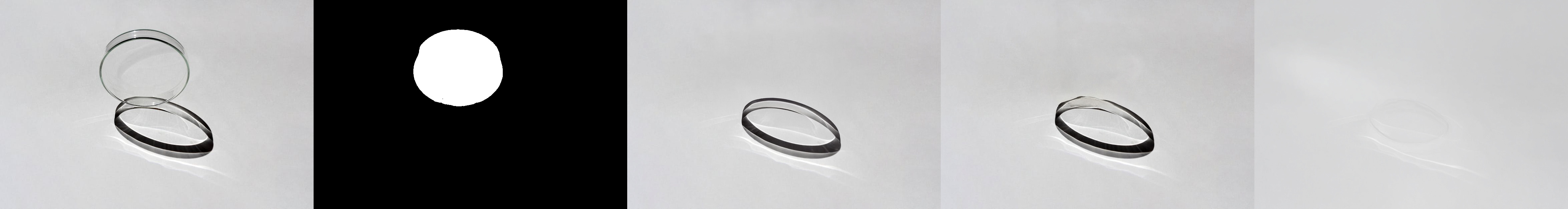}
    \par\vspace{0.05em}
    \includegraphics[width=0.92\textwidth]{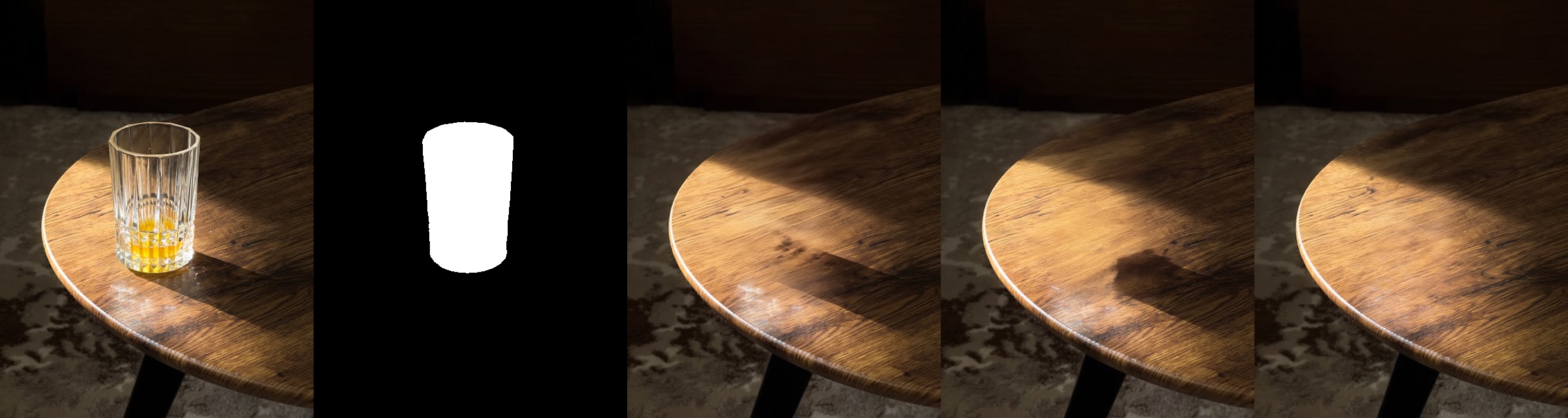}
    \par\vspace{0.05em}
    \includegraphics[width=0.92\textwidth]{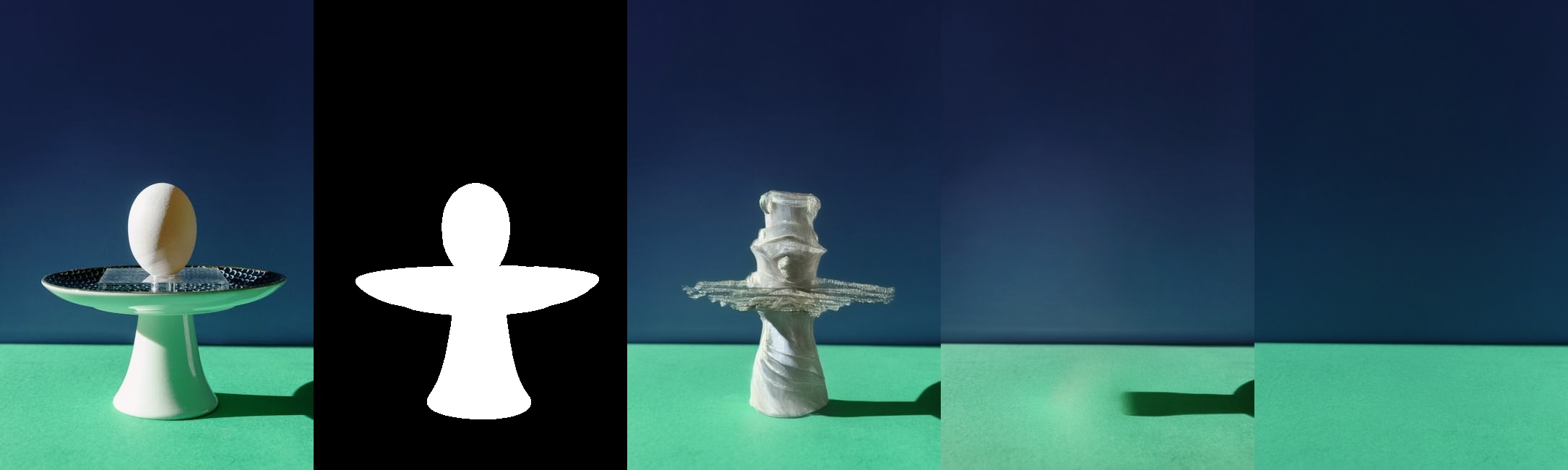}
    \par\vspace{0.05em}
    \includegraphics[width=0.92\textwidth]{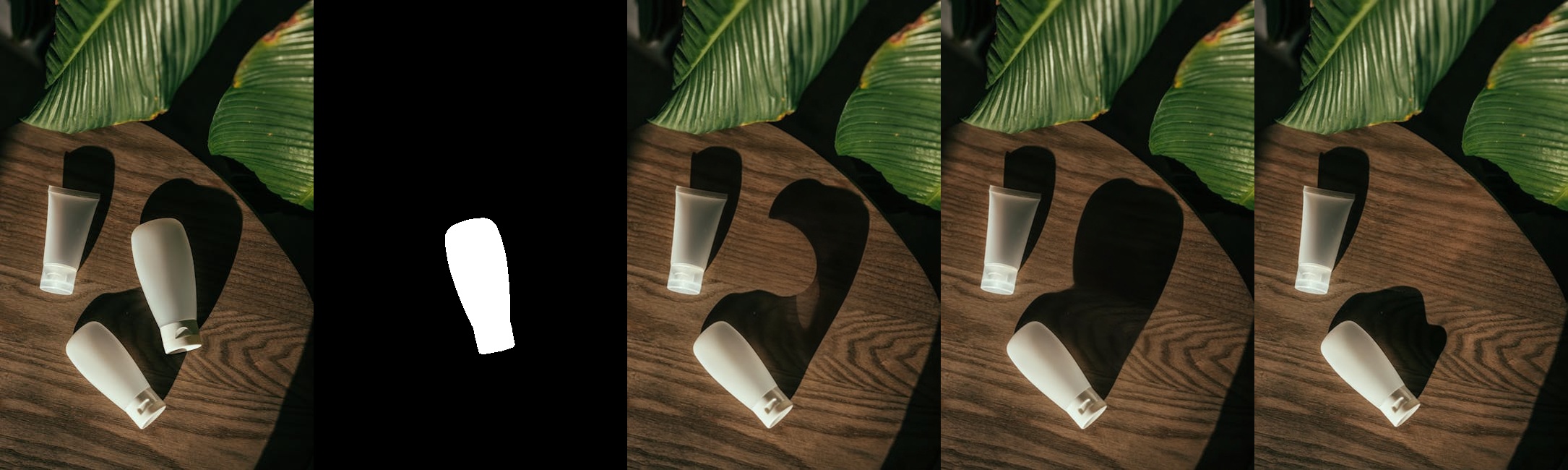}
    \caption{Nine additional hard-case qualitative examples from \textbf{Removal-HardEffects}, complementary to Figure~\ref{fig:removal-hardeffects-userstudy}. Each row shows one additional example. From left to right, columns show the original image, mask, \textbf{OpenData-30K}, \textbf{OpenData+SceneForge-30K}, and \textbf{SceneForge-16K}.}
    \label{fig:removal-hardeffects-supp-a}
\end{figure*}

Figure~\ref{fig:removal-hardeffects-supp-a} provides additional qualitative examples for the object-level removal study in Section~\ref{sec:exp-downstream}, complementing Figure~\ref{fig:removal-hardeffects-userstudy} in the main paper. Each row shows one additional hard case from \textbf{Removal-HardEffects} in the same visualization format as the main-paper figure. These examples further illustrate the importance of handling removal-associated shadows, reflections, and related local side effects.

\subsection*{Additional RemovalBench qualitative results}

\begin{figure*}[t]
    \centering
    \includegraphics[width=\textwidth]{figures/supp_bench/4_1x6.jpg}
    \par\vspace{0.15em}
    \includegraphics[width=\textwidth]{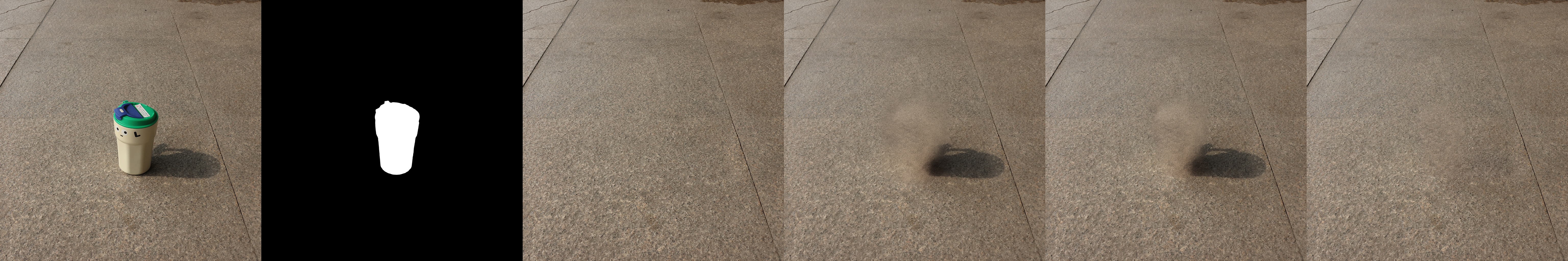}
    \par\vspace{0.15em}
    \includegraphics[width=\textwidth]{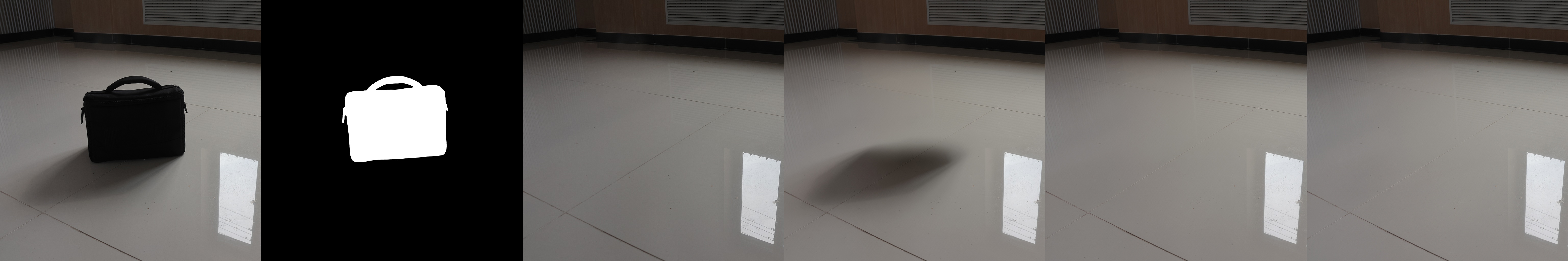}
    \par\vspace{0.15em}
    \includegraphics[width=\textwidth]{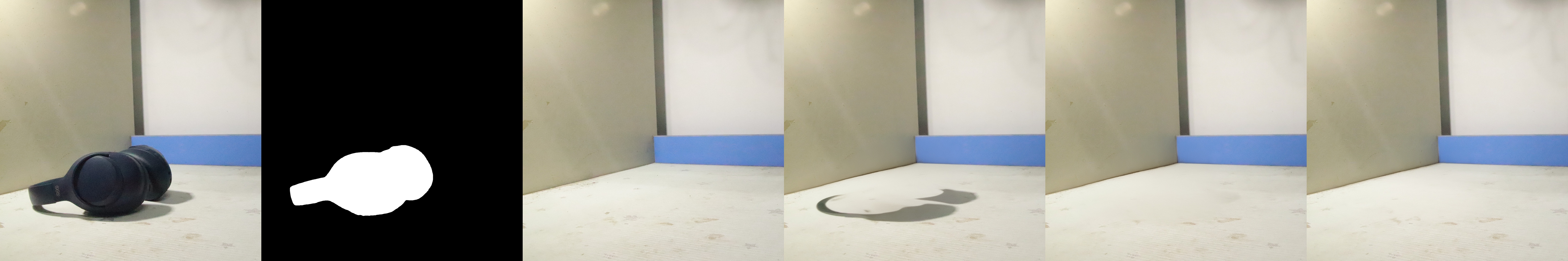}
    \par\vspace{0.15em}
    \includegraphics[width=\textwidth]{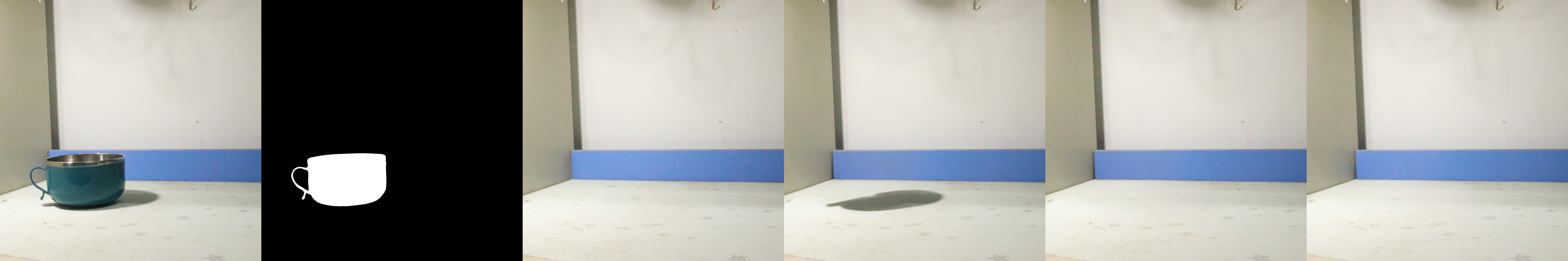}
    \par\vspace{0.15em}
    \includegraphics[width=\textwidth]{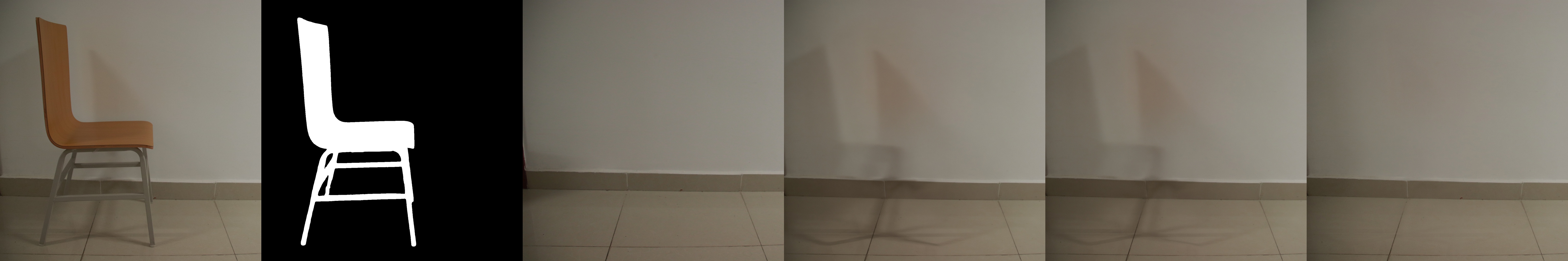}
    \par\vspace{0.15em}
    \includegraphics[width=\textwidth]{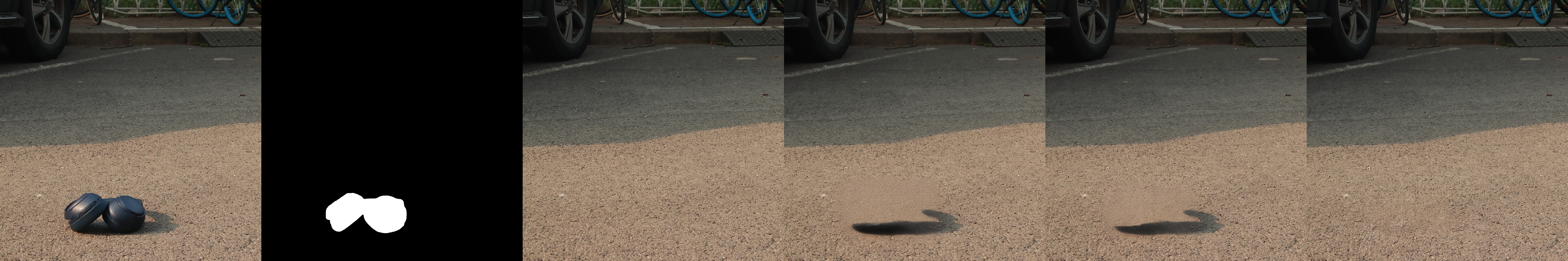}
    \par\vspace{0.15em}
    \includegraphics[width=\textwidth]{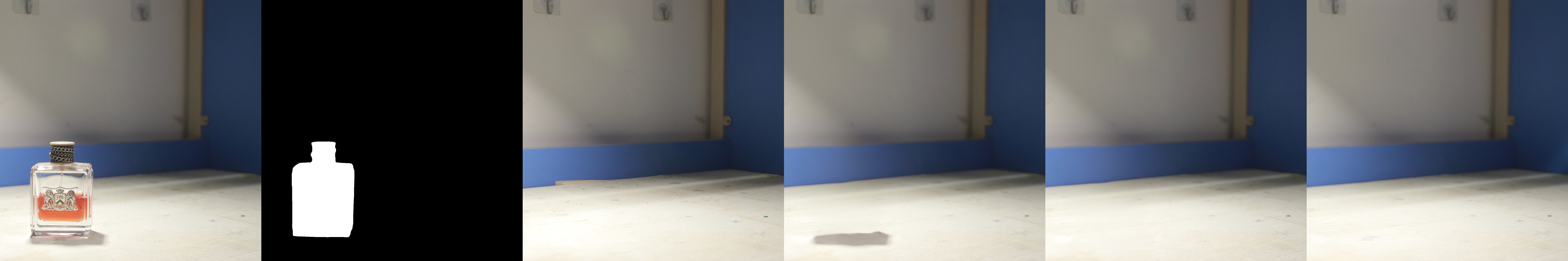}
    \par\vspace{0.15em}
    \includegraphics[width=\textwidth]{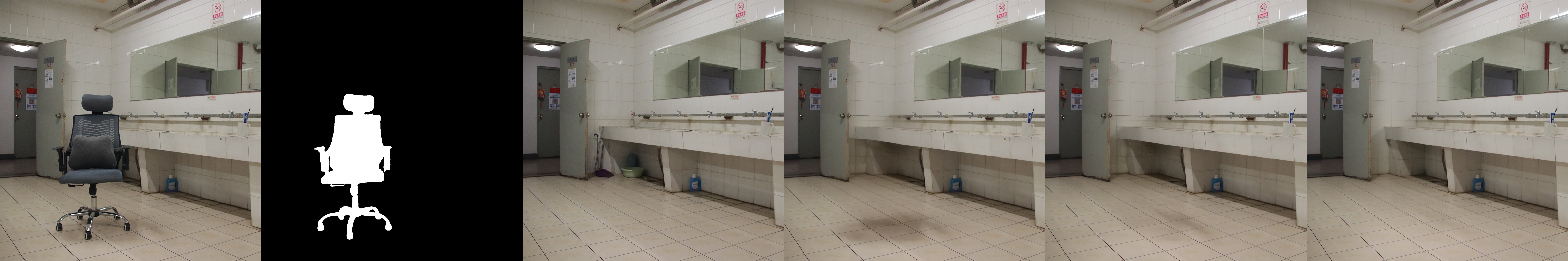}
    \caption{Additional \textbf{RemovalBench} qualitative results. Each row shows one example at full width. From left to right, columns show the original image, mask, ground-truth target, \textbf{OpenData-30K}, \textbf{OpenData+SceneForge-30K}, and \textbf{SceneForge-16K}.}
    \label{fig:removalbench-supp}
\end{figure*}

Figure~\ref{fig:removalbench-supp} shows additional qualitative results on the \textbf{RemovalBench} test set. Each row shows one example at full width. From left to right, columns show the original image, mask, ground-truth target, \textbf{OpenData-30K}, \textbf{OpenData+SceneForge-30K}, and \textbf{SceneForge-16K}.

\clearpage

\end{document}